\documentclass[sigconf, screen]{acmart}

\AtBeginDocument{
  }

\setcopyright{acmlicensed}
\acmConference{}{}{}
\copyrightyear{2026}
\acmYear{2026}
\acmDOI{XXXXXXX.XXXXXXX}
\acmISBN{978-1-4503-XXXX-X/26/10}
\usepackage[table]{xcolor}
\usepackage{bm}
\usepackage{algorithm}
\usepackage{algpseudocode}
\usepackage{booktabs}
\usepackage{multirow}
\usepackage{graphicx}
\usepackage{subcaption}
\usepackage{caption}
\usepackage{xcolor}
\usepackage{subcaption} % \begin{subfigure}
\acmSubmissionID{4461}

\setcopyright{none}           % 不显示版权信息

\acmConference[]{}{}{}        % 清空会议名
\acmDOI{}                     % 清空 DOI
\acmISBN{}                    % 清空 ISBN
\acmPrice{}                   % 清空价格
\acmYear{}                    % 清空年份
\copyrightyear{}              % 清空版权年份
\begin{document}

%%
%% The "title" command has an optional parameter,
%% allowing the author to define a "short title" to be used in page headers.
\title[ActDistill: General Action-Guided Self-Derived Distillation]{ActDistill: General Action-Guided Self-Derived Distillation \\ for Efficient Vision-Language-Action Models}

%%
%% The "author" command and its associated commands are used to define
%% the authors and their affiliations.
%% Of note is the shared affiliation of the first two authors, and the
%% "authornote" and "authornotemark" commands
%% used to denote shared contribution to the research.
\author{Wencheng Ye}
\authornote{Equal contribution.}
\affiliation{%
 \institution{School of Computer Science and Technology, Tongji University}
  \city{Shanghai}
  \country{China}
}
\email{2350227@tongji.edu.cn}

\author{Tianshi Wang}
\authornotemark[1]
\affiliation{%
 \institution{School of Computer Science and Technology, Tongji University}
  \city{Shanghai}
  \country{China}
}
\email{tswang0116@163.com}

\author{Lei Zhu}
\affiliation{%
  \institution{School of Computer Science and Technology, Tongji University}
  \city{Shanghai}
  \country{China}
}
\email{leizhu0608@gmail.com}

\author{Fengling Li}
\affiliation{%
  \institution{University of Technology Sydney}
  \country{Sydney, Australia}
}
\email{fenglingli2023@gmail.com}

\author{Guoli Yang}
\affiliation{%
  \institution{Advanced Institute of Big Data}
  \country{Beijing, China}
}
\email{yanggl@aibd.ac.cn}

\author{Heng Tao Shen}
\affiliation{%
  \institution{School of Computer Science and Technology, Tongji University}
  \city{Shanghai}
  \country{China}
}
\email{shenhengtao@hotmail.com}
%%
%% By default, the full list of authors will be used in the page
%% headers. Often, this list is too long, and will overlap
%% other information printed in the page headers. This command allows
%% the author to define a more concise list
%% of authors' names for this purpose.
\renewcommand{\shortauthors}{Trovato et al.}

%%
%% The abstract is a short summary of the work to be presented in the
%% article.
\begin{abstract}
Recent Vision-Language-Action (VLA) models have shown impressive flexibility and generalization, yet their practical deployment remains limited by substantial computational overhead and inference latency. In this work, we present ActDistill, a general action-guided self-derived distillation framework that transfers the action prediction capability of a full-scale VLA model to a lightweight counterpart. Unlike previous efficiency approaches that primarily focus on vision-language correlations, ActDistill explicitly leverages action priors to guide knowledge transfer and model compression, achieving action-oriented efficiency for VLA models. Specifically, we employ a well-trained VLA model as the teacher and introduce a graph-structured encapsulation to model the hierarchical dependencies in action prediction. The student model, derived from the graph-encapsulated teacher, is further equipped with a dynamic router that adaptively selects computation paths conditioned on action requirements. The routing process is guided by hierarchical graph-informed supervision, enabling efficient and stable action prediction. During inference, all auxiliary components are removed, allowing the student to execute only dynamically routed layers and generate high-precision actions. Experiments on embodied benchmarks demonstrate that ActDistill achieves comparable or superior performance to full-scale VLA models while reducing computation by over 50\% with up to 1.67$\times$ speedup, establishing a general paradigm for efficient embodied intelligence.
\end{abstract}


\maketitle

\section{Introduction}
\label{sec:intro}
Recent advances in Vision-Language Models (VLMs) \cite{VLM_Survey, Flamingo, OpenFlamingo} have enabled Vision-Language-Action (VLA) models to comprehend complex scenes, interpret task instructions, and generate continuous, executable action sequences \cite{VLA_Survey2,li2024visionlanguagefoundationmodelseffective,RoboMamba}. These models realize end-to-end intelligence that integrates perception, cognition, decision, planning, and execution, and have been widely adopted in robotic manipulation, visual navigation, and interactive control, forming a cornerstone of general embodied intelligence \cite{VLA_Survey1}.

Despite their strong multimodal reasoning and action-prediction capabilities, deploying VLA models in robotics remains challenging due to substantial computational cost and latency. Their large architectures, frequent cross-modal interactions, and complex action decoding substantially increase processing overhead, constraining their use in real-time or resource-limited settings \cite{EVLA_Survey}. Enabling efficient reasoning while preserving task performance is therefore critical for practical embodied applications \cite{yu2026surveyefficientvisionlanguageactionmodels}.

\begin{figure}[!b]
\centering
\includegraphics[width=0.98\linewidth]{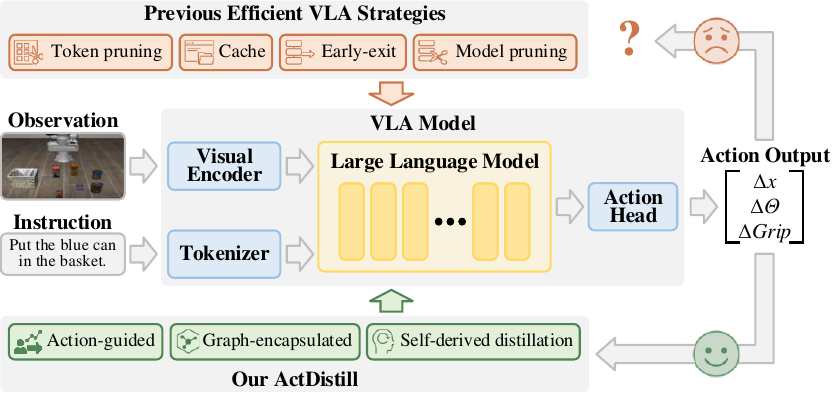}
\caption{Comparison between previous efficient VLA strategies and our proposed ActDistill.}
\label{fig:diagram}
\end{figure}

Researchers have explored various efficiency strategies for VLA models, including token pruning \cite{VLA-Cache, EfficientVLA}, early-exit \cite{DeeR-VLA, CEED-VLA}, and lightweight architectural design \cite{EdgeVLA, MoLe-VLA, vlaadapter}. While these approaches reduce computational cost to some extent, they largely inherit design principles from Vision-Language Models and overlook a fundamental property unique to VLA systems. As illustrated in Figure ~\ref{fig:diagram}, unlike pure VLMs whose representations are directly optimized for language generation, VLA models must transform rich vision-language features into compact, executable action signals. This transformation is not instantaneous, but occurs progressively across layers, where redundant perceptual and linguistic details are distilled into action-relevant representations \cite{zhang2026vlm4vlarevisitingvisionlanguagemodelsvisionlanguageaction}. 

However, existing methods do not account for this transformation process. Most approaches perform compression based on representational redundancy or temporal heuristics, without preserving the layers that are critical for the Vision-Language-to-Action (VL-to-Action) transformation. Even methods such as DeeR-VLA \cite{DeeR-VLA}, which incorporate action prediction into early-exit criteria, treat action quality as an output-level signal and ignore intermediate transformation dynamics. This limitation leads to two critical issues: \textit{1) critical information attrition}, where perceptual and semantic cues essential for grounding actions are inadvertently discarded, and \textit{2) action-semantic discontinuity}, where disrupting intermediate transformations weakens the consistency between perception and control, ultimately degrading policy stability.

To address the above limitations, we introduce ActDistill, a general action-guided distillation framework for efficient VLA models. The core idea is to explicitly align model compression with the VL-to-Action transformation process by leveraging action priors as the guiding signal. Specifically, we employ a well-trained VLA model as the teacher and propose a graph-structured encapsulation to model the hierarchical dependencies in action prediction. This design explicitly preserves the intermediate transformation structure from vision-language representations to executable actions, mitigating action-semantic discontinuity and ensuring stable semantic propagation across layers. Based on the graph-encapsulated teacher, we further derive a student model equipped with a dynamic routing mechanism, which adaptively selects computation layers conditioned on action requirements. By prioritizing layers that are critical for action prediction, the router avoids unnecessary computation while preventing the loss of task-relevant information. During inference, ActDistill executes only the dynamically selected key layers, significantly reducing computational cost while maintaining accurate and stable action prediction. In summary, our contributions are threefold:
\begin{itemize}
\item We propose ActDistill, a general action-guided distillation framework that explicitly aligns model compression with the VL-to-Action transformation process, enabling action-oriented efficiency in VLA models.
\item Technically, we introduce a graph-structured encapsulation to preserve the hierarchical transformation from vision-language representations to actions, along with an action-guided dynamic routing mechanism that selectively activates computation paths critical for action prediction.
\item Comprehensive experiments on representative VLA models and embodied benchmarks demonstrate that our ActDistill maintains task performance while reducing FLOPs by over 50\% and achieving up to 1.67$\times$ speedup.
\end{itemize}

\section{Related Work}
\label{sec:related_work}

\textbf{Vision-Language-Action Models.} 
The emergence of large-scale embodied datasets such as Open X-Embodiment \cite{OXE} has enabled the training of generalist robot policies across diverse embodiments, skills, and environments, laying the groundwork for VLA models. Building on these resources, \textit{autoregressive VLA models} such as OpenVLA \cite{OpenVLA}, SpatialVLA \cite{Spatialvla}, and WorldVLA \cite{WorldVLA} integrate visual perception, language understanding, and action prediction within a token-based sequence modeling framework, allowing robots to reason and act in the same manner as language models. In parallel, \textit{diffusion-based VLA models} such as Pi0 \cite{Pi0}, Pi0.5\cite{Pi05}, CogACT \cite{CogAct}, and GR00T N1 \cite{GR00T} formulate action prediction as conditional denoising or flow-matching processes, capturing multimodal action distributions and improving stability in high-dimensional control. Beyond imitation learning, \textit{reinforcement learning-augmented VLA models}, including VLA-RL \cite{VLA-RL} and SimpleVLA-RL \cite{SimpleVLA-RL}, demonstrate that online optimization can further enhance long-horizon planning and real-world adaptability. Despite differences in modeling paradigms, all these approaches share a common requirement: transforming high-dimensional vision-language representations into executable action signals. This transformation process \cite{OpenVLA, Pi0} introduces substantial computational overhead, making efficiency a critical challenge for real-world deployment.

\vspace{1.5mm}
\noindent\textbf{Efficiency Optimization for VLA Models.} 
To address the high computational cost and latency of VLA models, recent studies have explored diverse strategies for improving efficiency. \textit{Training-free approaches} enhance inference speed without additional optimization by exploiting redundancy reduction and caching. Methods such as VLA-Cache \cite{VLA-Cache}, FlashVLA \cite{FlashVLA}, and EfficientVLA \cite{EfficientVLA} reuse static visual tokens, skip redundant Transformer layers, or cache intermediate features, achieving notable acceleration with minimal accuracy degradation. \textit{Lightweight fine-tuning approaches} adapt well-trained VLA models for efficient execution. Frameworks such as DeeR-VLA \cite{DeeR-VLA}, MoLe-VLA \cite{MoLe-VLA}, and LightVLA \cite{LightVLA} introduce early-exit, dynamic activation, and differentiable token pruning to balance precision and speed through lightweight post-training. Beyond these adaptations, \textit{lightweight redesign approaches}, exemplified by TinyVLA \cite{TinyVLA}, SmolVLA \cite{SmolVLA}, and EdgeVLA \cite{EdgeVLA}, rebuild the model from scratch using compact backbones, non-autoregressive decoding, or quantized modules, ensuring deployability on resource-limited platforms. However, these approaches \cite{VLA-Cache, DeeR-VLA, TinyVLA} largely overlook the progressive transformation from vision-language representations to executable actions, and perform compression without preserving the layers critical for this process. In contrast, our ActDistill explicitly models the VL-to-Action transformation and leverages action guidance to preserve task-relevant computation during distillation and inference.

\begin{figure*}[t]
\centering
\includegraphics[width=0.983\linewidth]{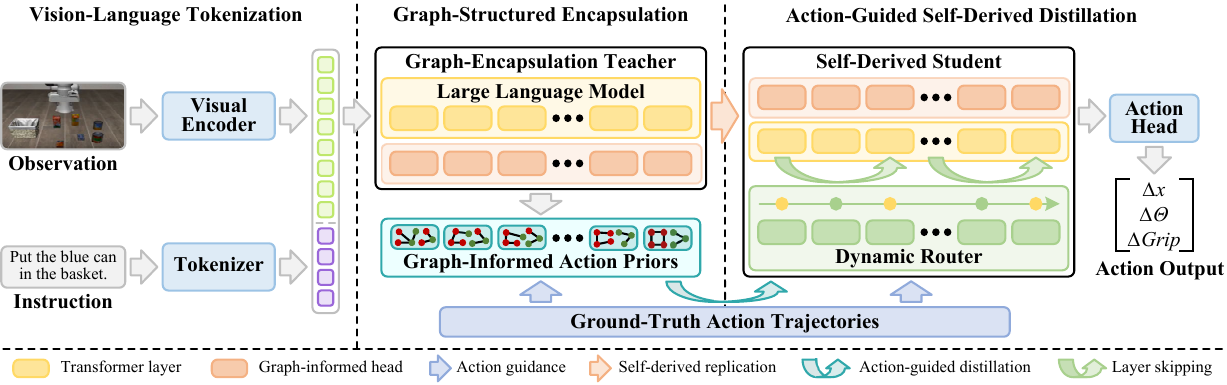}
\caption{Overview of our ActDistill framework.}
\label{fig:framework}
\end{figure*}

\section{ActDistill}
\label{sec:method}

\subsection{Problem Formulation}
We study embodied action prediction, where a Vision-Language-Action (VLA) model translates visual observations and linguistic instructions into executable robot actions. Given visual input $\bm{v}$ (e.g., RGB frames or multi-view perception) and a language instruction $\bm{l}$, the model outputs an action vector $\bm{a}$ encoding control parameters such as end-effector pose and gripper state. A typical VLA architecture includes a visual encoder $\mathcal{E}_v$, a language encoder $\mathcal{E}_l$, a multimodal backbone $\mathcal{B}$, and an action head $\mathcal{H}$. For an input pair $(\bm{v}, \bm{l})$, the encoders extract modality-specific features fused by $\mathcal{B}$, often implemented as Transformers with $L$ layers yielding intermediate states $\{\bm{h}_l\}_{l=1}^{L}$ that progress from spatial perception to task-level reasoning. The action head $\mathcal{H}$ converts the fused representation into executable control signals, using either an \textit{autoregressive} formulation $p(\bm{a}_t\mid\bm{v},\bm{l},\bm{a}_{<t})$ or a \textit{diffusion-based} formulation that synthesizes full trajectories via iterative denoising.

Although VLA models exhibit strong flexibility and generalization, their hierarchical representations contribute unevenly to action prediction. In particular, the backbone must transform rich vision-language features into executable action semantics across layers, yet only a subset of layers is critical for this Vision-Language-to-Action (VL-to-Action) transformation. Therefore, the goal of this work is to preserve these transformation-critical components while eliminating computation that does not contribute to action prediction. To this end, we develop a framework that captures action-centric semantics and selectively activates computation paths based on their relevance to action generation, enabling efficient yet faithful approximation of full-scale VLA models.

\subsection{Overview of ActDistill}
The core idea of ActDistill is to enable efficient embodied manipulation by transferring the action-centric semantics of a large VLA model to a lightweight student, as illustrated in Figure \ref{fig:framework}. Unlike conventional efficiency strategies that are driven by vision-language correlations and remain agnostic to the VL-to-Action transformation, ActDistill explicitly focuses on how multimodal representations are progressively converted into executable actions.

During efficiency optimization, the teacher VLA first performs graph-structured encapsulation, organizing intermediate representations into relational graphs, where nodes denote layer-wise features and edges capture their dependencies. From these graphs, the teacher derives action priors that characterize the contribution of intermediate representations to action prediction. The student model then aligns its representations with this hierarchical action-centric structure, focusing on task-relevant cues while discarding redundant correlations. In parallel, an action-guided dynamic router is co-optimized to adaptively select computation layers based on evolving action semantics, allocating computation to stages that are most critical for accurate action prediction.

During inference, the lightweight student executes only the dynamically selected layers, significantly reducing computational cost while maintaining accurate and stable action prediction. By integrating semantic encapsulation, action-guided distillation, and adaptive computation, our ActDistill provides a general and scalable framework for efficient VLA models.

\subsection{Graph-Structured Encapsulation}
\label{sec:graph}

\textbf{Motivation.} 
In VLA models, the intermediate visual-language representations often entangle spatial configurations and semantic dependencies, obscuring the components that are truly decisive for action prediction. To expose these dependencies, we reformulate the teacher's intermediate representations into graph-structured abstractions that capture the relational interplay among perceptual and linguistic cues. Whereas standard attention models dense global correlations, the graph formulation introduces explicit sparse structural priors that help separate manipulation-relevant interactions from irrelevant background signals. Critically, this graph is not a generic relational module: its topology is shaped end-to-end by action prediction objectives, ensuring that the learned sparse structure reflects true action-relevant dependencies rather than general perceptual correlations. This formulation establishes a structured bridge between perception, instruction, and control, providing a relational abstraction through which the student can acquire a more disentangled and action-centric understanding of the task.

\vspace{1.5mm}
\noindent\textbf{Graph-Structured Encapsulation Process.}
Given the $l$-th layer hidden representation $\bm{h}_l$ of the teacher backbone, each token feature $\bm{h}_{l,i}$ is treated as a node in a dynamic relational graph $\mathcal{G}_l=(\mathcal{V}_l,\mathcal{E}_l)$, where $\mathcal{V}_l$ denotes the node set and $\mathcal{E}_l$ represents learnable edges that encode contextual dependencies. 
The adjacency matrix $\hat{\bm{A}}_l$ is computed as
\begin{equation}
\hat{\bm{A}}_l(i,j) = \exp(\phi(\bm{h}_{l,i})^\top \psi(\bm{h}_{l,j})),
\end{equation}
where $\phi(\cdot)$ and $\psi(\cdot)$ are learnable linear projections that estimate attention-based affinities. 
Then, for each node $i$, only its $k$ largest affinity edges are retained to form a $k$-nearest neighbor graph, and the adjacency matrix is normalized:
\begin{equation}
\bm{A}_l(i,j) = \frac{\hat{\bm{A}}_l(i,j)}{\sum_{j \in \text{TopK}_k(i)} \hat{\bm{A}}_l(i,j)}, 
\quad j \in \text{TopK}_k(i),
\end{equation}
where $\text{TopK}_k(i)$ denotes the indices of the $k$ largest values in row $i$.

Each node then aggregates its neighborhood information via an attention-based message passing step over its $k$ nearest neighbors. We present a single-head formulation here:
\begin{equation}
\tilde{\bm{h}}_{l,i} = \sigma_r(\sum_{j \in \text{TopK}_k(i)} \bm{A}_l(i,j)\, \bm{W}_l \bm{h}_{l,j}),
\label{eq:graph_update_knn}
\end{equation}
where $\bm{W}_l$ is a layer-specific transformation matrix and $\sigma_r(\cdot)$ is a ReLU activation. By actively reasoning over these strictly relevant local neighbors via non-linear transformations, it ensures the updated features encode complex geometric dependencies rather than just aggregated signals. The updated representation $\tilde{\bm{h}}_l = \{\tilde{\bm{h}}_{l,i}\}_{i}$ preserves both local spatial context and high-level relational cues.

\vspace{1.5mm}
\noindent\textbf{Graph Learning in the Teacher.}
After graph construction, the teacher encodes relational semantics that directly support action generation. For the $l$-th layer, the updated node features $\tilde{\bm{h}}_{l}$ 
are aggregated into a structured semantic embedding via attention-based pooling:
\begin{equation}
\bm{s}_{l}^{\text{tea}} = \sum_{i=1}^{N} \alpha_{l,i}\,\tilde{\bm{h}}_{l,i},
\ \ \alpha_{l,i} = \frac{\exp(\bm{w}_{p}^{\top}\tilde{\bm{h}}_{l,i})}
{\sum_{j}\exp(\bm{w}_{p}^{\top}\tilde{\bm{h}}_{l,j})},
\label{eq:semantic_embedding_knn}
\end{equation}
where $N$ is the number of tokens in $\bm{h}_l$, and $\bm{w}_{p}$ is a learnable projection that measures each node's contribution to the final control decision. The resulting embedding $\bm{s}_{l}^{\text{tea}}$ acts as the teacher's structured semantic capsule, encoding spatial and linguistic relations essential for action prediction.

To align each capsule with executable actions, the teacher is trained with an auxiliary prediction loss:
\begin{equation}
\mathcal{L}_{\text{aux}}^{(l)} = \big\|\mathcal{H}_{l}^{\text{tea}}(\bm{s}_{l}^{\text{tea}}) - \bm{a} \big\|_{2}^{2},
\label{eq:teacher_action_loss_knn}
\end{equation}
where $\mathcal{H}_{l}^{\text{tea}}$ maps the semantic capsule to the ground-truth action $\bm{a}$. 
This auxiliary objective drives deeper layers to encode increasingly abstract yet action-consistent semantics.

Through this procedure, the teacher yields a hierarchy of compact and interpretable representations $\{\bm{s}_{l}^{\text{tea}}\}_{l=1}^{L}$ that isolate control-relevant cues from redundant correlations. These structured semantics later provide transferable supervision that enables the student to perform efficient and accurate embodied reasoning.

% Implementation note: in our experiments, we set k=8 for the k-nearest neighbor graph.

\subsection{Action-Guided Self-Derived Distillation}
\label{sec:distill}

\textbf{Motivation.}  
Although large-scale VLA models demonstrate strong multimodal reasoning, their dense architecture and redundant fusion operations hinder real-time robotic deployment. To preserve comparable action reasoning within efficiency constraints, we propose an action-guided self-derived distillation framework. Instead of compressing or pruning the teacher directly, the student reconstructs the teacher's decision-making process under the guidance of action priors. This self-derived paradigm encourages the student to rebuild hierarchical control reasoning through its own compact computation graph, selectively inheriting essential control cues while learning to decide which computations are necessary for accurate execution.

\vspace{1.5mm}
\noindent\textbf{Self-Derived Lightweight Replica.}
The student serves as a structurally aligned yet parameter-efficient counterpart to the teacher. It preserves the hierarchical organization while adopting a reduced depth to enable lightweight computation. To further support adaptive execution, a dynamic router $\mathcal{R}$ is introduced to determine, for each input, which subset of layers should be executed and which can be skipped.

The router models action-oriented cross-modal dependencies and predicts layer-wise gating scores. Given the visual embedding $\bm{v}$ and language embedding $\bm{l}$, the gating score of the $l$-th layer is defined as
\begin{equation}
g_{l} = \sigma_s(\bm{w}_{r,l}^{\top}[\bm{v}; \bm{l}]),
\label{eq:routing_gate}
\end{equation}
where $g_l \in [0,1]$ denotes the activation confidence of the $l$-th layer, $\bm{w}_{r,l}$ is a learnable layer-specific projection vector, and $\sigma_s(\cdot)$ is the sigmoid function. During training, $\mathcal{R}$ is initialized with low activation scores and jointly optimized with the distillation objective through soft gating, allowing the router to learn which layers are critical for accurate control. At inference time, the continuous scores are  discretized with a threshold $\tau$: layers with $g_l < \tau$ are skipped, and those with $g_l \geq \tau$ are executed. 
Through this mechanism, the student becomes a self-derived lightweight replica of the teacher, executing only action-relevant computations to achieve balance between accuracy and efficiency.

\vspace{1.5mm}
\noindent\textbf{Action-Guided Distillation Learning.} 
To ensure that the student's semantics align with executable control behaviors, we define an action-guided distillation objective that combines semantic alignment and action consistency.  
For each layer, the semantic loss enforces both instance-level alignment and relational preservation:
\begin{equation}
\begin{aligned}
\mathcal{L}_{\text{sem}}^{(l)} =&
\mathbb{E}[1-\mathrm{sim}(\bm{s}_{l}^{\text{stu}},\bm{s}_{l}^{\text{tea}})]
\\+& \eta\,\|\mathrm{Sim}(\bm{s}_{l}^{\text{stu}})-\mathrm{Sim}(\bm{s}_{l}^{\text{tea}})\|_{F}^{2},
\label{eq:semantic_loss}
\end{aligned}
\end{equation}
where $\mathrm{sim}(\cdot,\cdot)$ denotes the cosine similarity between sample pairs, and $\mathrm{Sim}(\cdot)$ measures relational similarity at the batch level. To ensure precise and stable control, we introduce a triple-MSE loss for action prediction:
\begin{equation}
\begin{aligned}
\mathcal{L}_{\text{act}}^{(l)} &= 
\|\mathcal{H}_{l}^{\text{stu}}(\bm{s}_{l}^{\text{stu}})-\bm{a}\|_{2}^{2} 
\\&+ \|\mathcal{H}_{l}^{\text{stu}}(\bm{s}_{l}^{\text{stu}})-\mathcal{H}_{l}^{\text{tea}}(\bm{s}_{l}^{\text{tea}})\|_{2}^{2} 
\\&+ \|\mathcal{H}_{l}^{\text{stu}}(\bm{s}_{l}^{\text{stu}})-\mathrm{sg}(\mathcal{H}_{l-1}^{\text{stu}}(\bm{s}_{l-1}^{\text{stu}}))\|_{2}^{2},
\label{eq:action_loss}
\end{aligned}
\end{equation}
where $\mathrm{sg}(\cdot)$ denotes the stop-gradient operation, which blocks backpropagation to shallower layers and promotes progressive refinement from perception to control. For the first layer, the third term is omitted.

Together, these objectives ensure that the student aligns with the teacher's structured semantics while maintaining accurate and stable control.

\subsection{Training and Inference}
\label{sec:train_infer}
The overall training and inference procedure of ActDistill is outlined in Algorithm~\ref{alg:actdistill}, and this section provides a detailed description of the corresponding procedures.

\vspace{1.5mm}
\noindent\textbf{Training Objective.}
The overall objective integrates semantic alignment, action consistency, and routing regularization into a unified optimization framework. Given the per-layer semantic loss $\mathcal{L}_{\text{sem}}^{(l)}$ and action loss $\mathcal{L}_{\text{act}}^{(l)}$, the overall distillation objective is formulated as
\begin{equation}
\mathcal{L}_{\text{distill}} =\sum_{l=1}^{L}\lambda_{l}(\alpha\,\mathcal{L}_{\text{sem}}^{(l)}+\beta\,\mathcal{L}_{\text{act}}^{(l)}),
\label{eq:distillation_total}
\end{equation}
where $\lambda_{l}={(l/L)^{2}}$ assigns higher weights to deeper layers to reflect their greater role in high-level decision reasoning, and $\alpha$, $\beta$ balance the semantic and action components. 

To encourage efficient routing behavior, we introduce an additional load-balancing term $\mathcal{L}_{\text{lb}}$ to regulate the distribution of activated layers:
\begin{equation}
\mathcal{L}_{\text{lb}} = \sum_{l=1}^{L}(g_{l}-\bar{g})^{2}, \ \ \ \bar{g} = \frac{1}{L}\sum_{l}g_{l},
\end{equation}
where $g_{l}$ is the routing gate from Eq.~(\ref{eq:routing_gate}).  
The final training loss combines these terms:
\begin{equation}
\mathcal{L}_{\text{total}} = \mathcal{L}_{\text{distill}} + \gamma\,\mathcal{L}_{\text{lb}},
\label{eq:training_total}
\end{equation}
with $\gamma$ controlling the regularization strength.  
The teacher model is frozen during training, and all supervision signals are derived from its pre-computed structured semantics and action predictions. The student, router, and projection heads are optimized jointly in an end-to-end fashion. 

\begin{algorithm}[t]
\caption{\ Training and Inference of ActDistill}
\label{alg:actdistill}
\small
\begin{algorithmic}[1]
\Require Teacher $\mathcal{T}$, data $\mathcal{D}=\{(\bm{v},\bm{l},\bm{a})\}$ 
%\Ensure Student $\mathcal{S}$ and router $\mathcal{R}$
\vspace{3pt}
\Statex \textbf{Training I: Graph-Structured Encapsulation}
\State Freeze $\mathcal{T}$, initialize graphs $\{\mathcal{G}_l\}$ and heads $\{\mathcal{H}_l^{\text{tea}}\}$
\For{each mini-batch $(\bm{v},\bm{l},\bm{a}) \in \mathcal{D}$}
    \State Extract teacher features $\{\bm{h}_l^{\text{tea}}\}\gets\mathcal{T}(\bm{v},\bm{l})$
    \State Build graphs $\mathcal{G}_l(\bm{h}_l^{\text{tea}})$ and predict $\hat{\bm{a}}_l^{\text{tea}} \gets \mathcal{H}_l^{\text{tea}}(\bm{s}_l^{\text{tea}})$
    \State Update $\{\mathcal{G}_l,\mathcal{H}_l^{\text{tea}}\}$ via Eq.(\ref{eq:teacher_action_loss_knn})
\EndFor
\vspace{3pt}
\Statex \textbf{Training II: Action-Guided Self-Derived Distillation}
\State Initialize $\mathcal{S}$, router $\mathcal{R}$, and heads $\{\mathcal{H}_l^{\text{stu}}\}$
\For{each mini-batch $(\bm{v},\bm{l},\bm{a}) \in \mathcal{D}$}
    \State Compute routing gates $\{g_l\} \gets \mathcal{R}(\mathcal{E}_v(\bm{v}),\mathcal{E}_l(\bm{l}))$
    \State Run layers with soft gates $\{g_l\}$ to obtain $\{\bm{s}_l^{\text{stu}}, \hat{\bm{a}}_l^{\text{stu}}\}_{l=1}^L$

    \State Update $\mathcal{S}$ and $\mathcal{R}$ via Eq.(\ref{eq:training_total})
\EndFor
\vspace{3pt}
\Statex \textbf{Inference}
\State $g_l\!\leftarrow\!\mathcal{R}(E_v(\bm{v}),E_l(\bm{l}))$, execute layers with $g_l\!\ge\!\tau$
\State Output $\hat{\bm{a}}=\mathcal{H}^{\text{stu}}(\bm{s}_{L^*}^{\text{stu}})$
\end{algorithmic}
\end{algorithm}

\vspace{1.5mm}
\noindent\textbf{Inference Procedure.}
At inference time, only the student and the trained router $\mathcal{R}$ are retained.  
Given an input pair $(\bm{v},\bm{l})$, the router first produces layer-wise activation scores $\{g_{l}\}_{l=1}^{L}$.  
Each layer is executed if $g_{l}\geq\tau$, where $\tau$ is a pre-defined threshold controlling the efficiency-accuracy trade-off.  
This dynamic execution produces an adaptive computation path tailored to the input complexity.  
The final control command is generated as
\begin{equation}
\hat{\bm{a}} = \mathcal{H}^{\text{stu}}(\bm{s}_{L^{*}}^{\text{stu}}),
\end{equation}
where $L^{*}$ is the last executed layer under routing, and $\mathcal{H}^{\text{stu}}$ produces the corresponding action vector. With this selective inference, ActDistill enables real-time action generation with markedly lower latency and computational cost.

\vspace{1.5mm}
\noindent\textbf{Discussion.}
The unified training–inference design enables ActDistill to effectively balance efficiency and accuracy within a single learning paradigm. During training, semantic alignment enhances control precision, while action-guided gradients steer the router toward computation paths critical for action prediction. During inference, the router adaptively allocates computation based on the distilled semantics, preserving the fidelity of the teacher’s reasoning under real-time constraints. Moreover, as ActDistill operates solely on the multimodal backbone, it is compatible with both autoregressive and diffusion-based VLA models. This synergy allows ActDistill to maintain strong embodied reasoning across diverse hardware platforms and model architectures, providing a practical and scalable solution for efficient embodied intelligence.

\section{Experiment}
\label{sec:experiment}

\subsection{Experimental Setup}
\label{sec:experimental_setup}

\noindent\textbf{Backbones.}
We evaluate ActDistill on two representative VLA models: OpenVLA~\cite{OpenVLA} and CogACT~\cite{CogAct}. OpenVLA follows an autoregressive paradigm that predicts action tokens sequentially from visual and linguistic inputs while CogACT adopts a diffusion-based paradigm that produces continuous trajectories through iterative denoising to capture smooth transitions and uncertainty. These two backbones therefore provide complementary action-prediction settings for evaluating ActDistill across heterogeneous architectures.

\vspace{1.5mm}
\noindent\textbf{Benchmarks.}
Experiments are conducted on two embodied manipulation benchmarks, LIBERO \cite{Libero} and SIMPLER \cite{SIMPLER}. LIBERO consists of four task suites, spatial, object, goal, and long, covering spatial reasoning, object-centric manipulation, goal-conditioned control, and long-horizon planning. SIMPLER consists of two scenarios, Visual Matching and Variant Aggregation, where the former focuses on grounding actions through visual correspondences and the latter evaluates generalization across object and scene variations. Each scenario includes four tasks: pick coke can, move near, open/close drawer, and open top drawer and place apple. Following native settings, OpenVLA is evaluated on LIBERO and CogACT on SIMPLER, consistent with their original training benchmarks.

\vspace{1.5mm}
\noindent\textbf{Baselines.}
We compare with five state-of-the-art efficient VLA methods. VLA-Cache \cite{VLA-Cache} reuses visual tokens across frames to reduce redundant computation. EfficientVLA \cite{EfficientVLA} accelerates inference by pruning redundant layers and visual tokens. SparseVLM \cite{SparseVLM} sparsifies visual features to cut unnecessary processing. FastV \cite{FastV} simplifies visual-language fusion for speed-oriented deployment. MoLe-VLA \cite{MoLe-VLA} introduces dynamic layer routing with cognitive distillation to balance accuracy and efficiency. 
These methods represent complementary efficiency paradigms in caching, pruning, sparsification, and adaptive routing, and are generally evaluated on the corresponding benchmarks of their backbones.

\vspace{1.5mm}
\noindent\textbf{Implementation Details.}
Our ActDistill is trained on Open X-Embodiment's \cite{OXE} Berkeley Bridge subset using 4 NVIDIA RTX 5090 GPUs for approximately 8 hours. We use AdamW with a learning rate of $1\times10^{-6}$, cosine annealing, and a weight decay of $0.01$ and a batch size of 128. Key hyper-parameters are set to $\alpha\!=\!1$, $\beta\!=\!1$, $\eta\!=\!0.5$ and load-balancing loss term $\gamma=\!0.05$. The graph-encapsulation module uses a two-layer Graph ATtention network (GAT) \cite{velickovic2018graph} with $k = 8$ neighbors and attention pooling, and the routing threshold $\tau=0.5$ is used. 

\begin{table*}[!t]
\centering
\caption{Performance comparison on the LIBERO benchmark using the OpenVLA backbone.}
\vspace{-1mm}
\label{tab:libero}
\renewcommand{\arraystretch}{0.975}
\setlength{\tabcolsep}{10pt}
\resizebox{\linewidth}{!}{
\begin{tabular}{lcccccccc}
\toprule
\multirow{2}{*}{Methods} & \multirow{2}{*}{Sources} & \multicolumn{5}{c}{Success Rate ($\uparrow$)} & \multirow{2}{*}{Speed-up ($\uparrow$)} & \multirow{2}{*}{FLOPs ($\downarrow$)} \\
\cmidrule(lr){3-7}
 & & Spatial & Object & Goal & Long & Average &  &  \\
\midrule
{\color{gray} OpenVLA \cite{OpenVLA}} & {\color{gray} Arxiv'24} & {\color{gray} 84.40\%} & {\color{gray} 86.60\%} & {\color{gray} 75.60\%} & {\color{gray} 53.20\%} & {\color{gray} 74.95\%} & {\color{gray} 1.00$\times$} & {\color{gray} 100.00\%} \\ 
FastV \cite{FastV} & ECCV'24 & 83.40\% & 84.00\% & 74.20\% & 51.60\% & 73.30\% & 1.11$\times$ & 41.60\% \\
SparseVLM \cite{SparseVLM} & ICML'25 & 79.80\% & 67.00\% & 72.60\% & 39.40\% & 64.70\% & 1.13$\times$ & 75.20\% \\
VLA-Cache \cite{VLA-Cache} & NeurIPS'25 & 83.80\% & 85.80\% & 76.40\% & 52.80\% & 74.70\% & 1.46$\times$ & 80.80\% \\
\textbf{ActDistill} & \textbf{Ours} & \textbf{81.80\%} & \textbf{85.80\%} & \textbf{72.40\%} & \textbf{55.80\%} & \textbf{73.95\%} & \textbf{1.59$\times$} & \textbf{49.50\%} \\
\bottomrule
\end{tabular}}
\end{table*}

\begin{table*}[t]
\centering
\caption{Performance comparison on the SIMPLER benchmark using the CogACT backbone under two scenarios, Visual Matching and Variant Aggregation.}
\vspace{-1mm}
\label{tab:simpler}
\renewcommand{\arraystretch}{0.975}
\setlength{\tabcolsep}{8.5pt}
\resizebox{\linewidth}{!}{
\begin{tabular}{lc@{\hskip 12pt}ccccccc}
\toprule
\multirow{2}{*}{Methods} & \multirow{2}{*}{Sources} & \multicolumn{5}{c}{Success Rate ($\uparrow$)} & \multirow{2}{*}{Speed-up ($\uparrow$)} & \multirow{2}{*}{FLOPs ($\downarrow$)} \\
\cmidrule(r){3-7}
 &  & PickCan & MoveNear & Drawer & DrawerApple & Average &  &  \\
\midrule
\rowcolor{gray!10}
\multicolumn{9}{l}{\textbf{Visual Matching}} \\
{\color{gray} CogACT~\cite{CogAct}} & {\color{gray} Arxiv'24} & {\color{gray} 91.30\%} & {\color{gray} 85.00\%} & {\color{gray} 71.80\%} & {\color{gray} 50.90\%} & {\color{gray} 74.75\%} & {\color{gray} 1.00$\times$} & {\color{gray} 100.00\%} \\
VLA-Cache~\cite{VLA-Cache} & NeurIPS'25 & 92.00\% & 83.30\% & 70.50\% & 51.60\% & 74.35\% & 1.47$\times$ & 80.10\% \\
EfficientVLA~\cite{EfficientVLA} & Arxiv'25 & 95.30\% & 82.40\% & 70.30\% & 56.50\% & 76.13\% & 1.53$\times$ & 44.10\% \\
MoLe-VLA~\cite{MoLe-VLA} & Arxiv'25 & 86.40\% & 80.20\% & 70.60\% & 50.40\% & 71.90\% & 1.58$\times$ & 46.20\% \\ 
\textbf{ActDistill} & \textbf{Ours} & \textbf{91.00\%} & \textbf{82.30\%} & \textbf{70.60\%} & \textbf{52.40\%} & \textbf{74.08\%} & \textbf{1.67$\times$} & \textbf{42.30\%} \\
\midrule
\rowcolor{gray!10}
\multicolumn{9}{l}{\textbf{Variant Aggregation}} \\
{\color{gray} CogACT~\cite{CogAct}} & {\color{gray} Arxiv'24} & {\color{gray} 89.60\%} & {\color{gray} 80.80\%} & {\color{gray} 28.30\%} & {\color{gray} 46.60\%} & {\color{gray} 61.33\%} & {\color{gray} 1.00$\times$} & {\color{gray} 100.00\%} \\
VLA-Cache~\cite{VLA-Cache} & NeurIPS'25 & 91.70\% & 79.30\% & 32.50\% & 45.80\% & 62.33\% & 1.37$\times$ & 80.10\% \\
EfficientVLA~\cite{EfficientVLA} & Arxiv'25 & 94.80\% & 77.60\% & 28.40\% & 51.90\% & 63.18\% & 1.54$\times$ & 45.10\% \\
MoLe-VLA~\cite{MoLe-VLA} & Arxiv'25 & 89.20\% & 79.50\% & 29.90\% & 46.20\% & 61.20\% & 1.56$\times$ & 43.20\% \\
\textbf{ActDistill} & \textbf{Ours} & \textbf{90.20\%} & \textbf{81.50\%} & \textbf{29.60\%} & \textbf{45.80\%} & \textbf{61.78\%} & \textbf{1.65$\times$} & \textbf{42.30\%} \\
\bottomrule
\end{tabular}}
\end{table*}

\begin{table*}[t]
\centering
\caption{Ablation study on the SIMPLER benchmark (Visual Matching scenario), analyzing the impact of structural design choices and loss components.}
\vspace{-1mm}
\label{tab:simpler_vm_all}
\renewcommand{\arraystretch}{0.975}
\setlength{\tabcolsep}{8.5pt}
\resizebox{\linewidth}{!}{
\begin{tabular}{lcccccccc}
\toprule
\multirow{2}{*}{Variants} & \multicolumn{5}{c}{Success Rate ($\uparrow$)} & \multirow{2}{*}{Speed-up ($\uparrow$)} & \multirow{2}{*}{FLOPs ($\downarrow$)} \\
\cmidrule(lr){2-6}
 & PickCan & MoveNear & Drawer & DrawerApple & Average &  &  \\
\midrule
{\color{gray} CogACT~\cite{CogAct}} & {\color{gray}91.30\%} & {\color{gray}85.00\%} & {\color{gray}71.80\%} & {\color{gray}50.90\%} & {\color{gray}74.75\%} & {\color{gray}1.00$\times$} & {\color{gray}100.0\%} \\
ActDistill & 91.00\% & 82.30\% & 70.60\% & 52.40\% & 74.08\% & 1.67$\times$ & 42.30\% \\ \midrule
Replace GAT with MLP & 89.20\% & 84.90\% & 45.20\% & 38.80\% & 64.53\% & 1.56$\times$ & 41.70\% \\
w/o semantic loss $\mathcal{L}_{\text{sem}}^{(l)}$ & 88.10\% & 80.50\% & 67.30\% & 43.70\% & 69.80\% & 1.60$\times$ & 42.50\% \\
w/o action loss $\mathcal{L}_{\text{act}}^{(l)}$ & 87.50\% & 79.10\% & 65.40\% & 44.10\% & 69.05\% & 1.63$\times$ & 41.90\% \\
w/o load-balancing $\mathcal{L}_{\text{lb}}$ & 87.50\% & 77.90\% & 62.80\% & 45.10\% & 68.33\% & 1.59$\times$ & 41.30\% \\ 
w/o relation-level semantic loss & 90.20\% & 81.30\% & 69.10\% & 50.40\% & 72.75\% & 1.59$\times$ & 42.20\% \\
\bottomrule
\end{tabular}}
\end{table*}

\subsection{Comparison with State-of-the-art}
\label{sec:comparison}

\noindent\textbf{Task Performance.}
Tables~\ref{tab:libero} and \ref{tab:simpler} present quantitative comparisons in terms of task success rate, speed improvement, and computational cost, evaluating both effectiveness and efficiency. 

Across all benchmarks, ActDistill delivers comparable or superior task success while maintaining strict efficiency constraints. 
As shown in Table~\ref{tab:libero}, it achieves an average success rate of 73.95\% on the LIBERO benchmark, only 1.0\% lower than the full OpenVLA model, and surpasses the baseline by 2.6\% on the long-horizon LIBERO-Long suite. 
This improvement highlights its robustness on tasks that demand extended temporal reasoning. 
A consistent trend is observed on the SIMPLER benchmark, as summarized in Table~\ref{tab:simpler}. 
In the Visual Matching scenario, ActDistill attains 74.08\% average success, nearly identical to the CogACT baseline with a marginal drop of 0.67\%. 
In the Variant Aggregation scenario, the performance slightly improves by 0.45\%. 
The largest gains occur on challenging tasks, with an increase of 1.5\% on DrawerApple in the Visual Matching scenario and 1.3\% on Drawer in the Variant Aggregation scenario. 
These results demonstrate that action-guided distillation not only preserves accuracy but also serves as an implicit regularization mechanism, enhancing stability on complex embodied control tasks.

\vspace{1.5mm}
\noindent\textbf{Efficiency.}
ActDistill consistently achieves substantial acceleration and computational savings across different architectures. On the LIBERO benchmark, as reported in Table~\ref{tab:libero}, it reduces inference latency from 48.91\,ms to 30.74\,ms, yielding a 1.59$\times$ speedup while reducing computation to only 49.5\% of the full OpenVLA model. On the SIMPLER benchmark, summarized in Table~\ref{tab:simpler}, ActDistill shortens inference time from 47.29\,ms to 28.31\,ms in the Visual Matching scenario and from 46.94\,ms to 28.42\,ms in the Variant Aggregation scenario, corresponding to 1.67$\times$ and 1.65$\times$ speedup, respectively, with over 55\% reduction in FLOPs. The slightly smaller improvement on OpenVLA can be attributed to its tightly coupled autoregressive structure, which limits flexible computation reduction, whereas CogACT benefits more from its decoupled cognitive-action design that better aligns with semantic distillation. Overall, ActDistill achieves around 40\% latency reduction and 50-60\% lower computational cost while maintaining control fidelity, demonstrating both its effectiveness and generality for efficient reasoning across diverse VLA architectures.

\begin{figure}[t]
\centering
\includegraphics[width=0.98\linewidth]{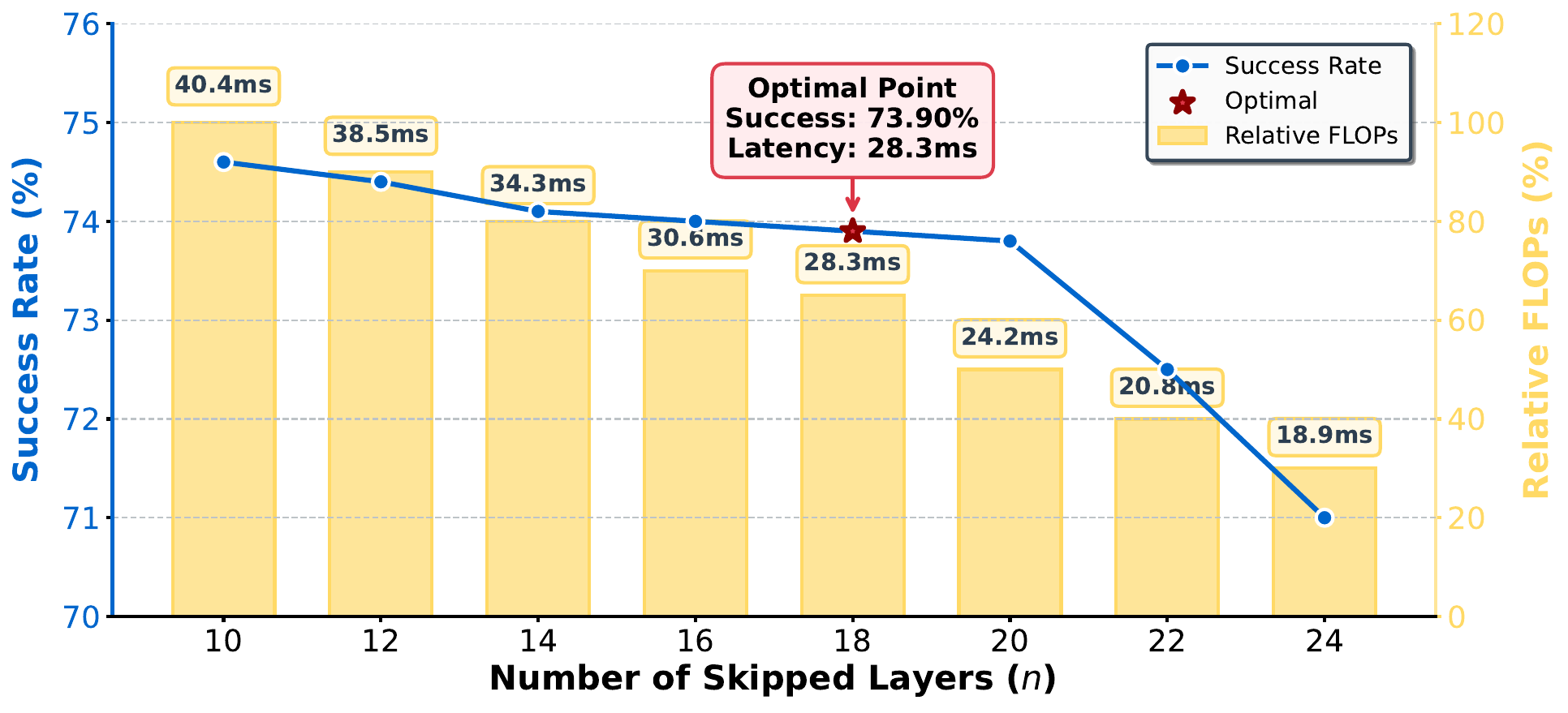}
\caption{Performance-efficiency trade-off across different layer skipping configurations.}
\label{fig:skip_curve}
\end{figure}

\begin{figure}[t]
\centering
\includegraphics[width=0.98\linewidth]{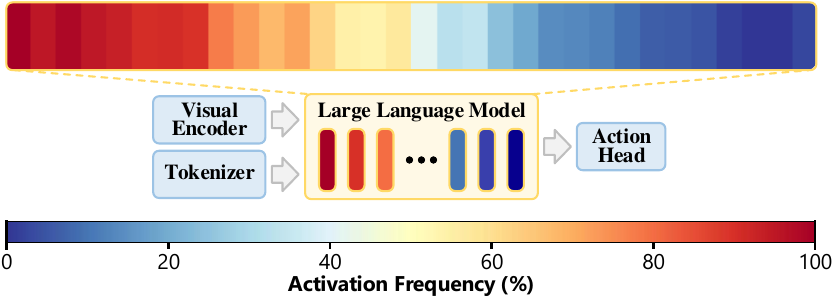}
\caption{Visualization of layer-wise activation frequency across the VLA backbone.}
\label{fig:visualization}
\end{figure}

\subsection{Ablation Study}
\label{sec:ablation}
We evaluate the influence of ActDistill's core components on the SIMPLER benchmark \cite{SIMPLER}, focusing on graph-structured encapsulation and loss design, with results shown in Table~\ref{tab:simpler_vm_all}. All ablations are evaluated under a similar computational budget, isolating the effect of each component.

\vspace{1.5mm}
\noindent\textbf{Graph-Structured Encapsulation.}
Replacing the GAT with an MLP causes the average success rate to drop from 74.08\% to 64.53\%, with the Drawer and DrawerApple tasks declining by 25.4\% and 13.6\%. This indicates that explicit relational modeling is crucial for capturing spatial dependencies in manipulation.

\vspace{1.5mm}
\noindent\textbf{Loss Components.}
Removing the semantic loss $\mathcal{L}_{\text{sem}}^{(l)}$ or action loss $\mathcal{L}_{\text{act}}^{(l)}$ lowers performance to 69.80\% and 69.05\%, indicating their complementarity. 
Disabling the load-balancing loss $\mathcal{L}_{\text{lb}}$ further reduces performance to 68.33\%, while removing the relation-level semantic loss yields 72.75\%. 
These results validate that multi-level supervision and relational consistency are key to maintaining control precision under efficiency constraints.

In summary, the graph-structured encapsulation together with dynamic routing sustain ActDistill's robust performance while enabling efficient embodied reasoning.

\begin{figure}[t]
    \centering
    \begin{subfigure}[t]{0.46\linewidth}
        \centering
        \includegraphics[width=\linewidth]{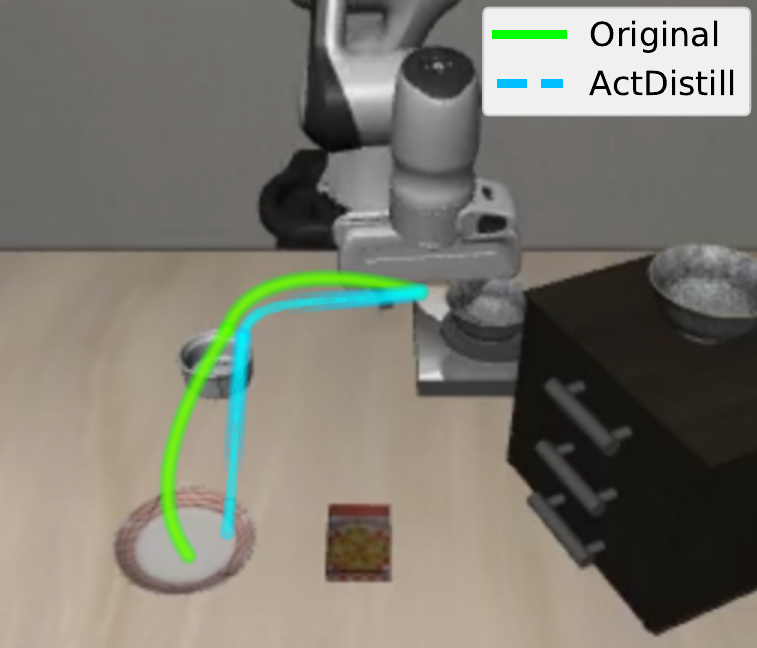}
        \caption{Instruction: Pick up the black bowl on the stove and place it on the plate.}
        \label{fig:traj_a}
    \end{subfigure}
    \hspace{0.03\linewidth}
    \begin{subfigure}[t]{0.46\linewidth}
        \centering
        \includegraphics[width=\linewidth]{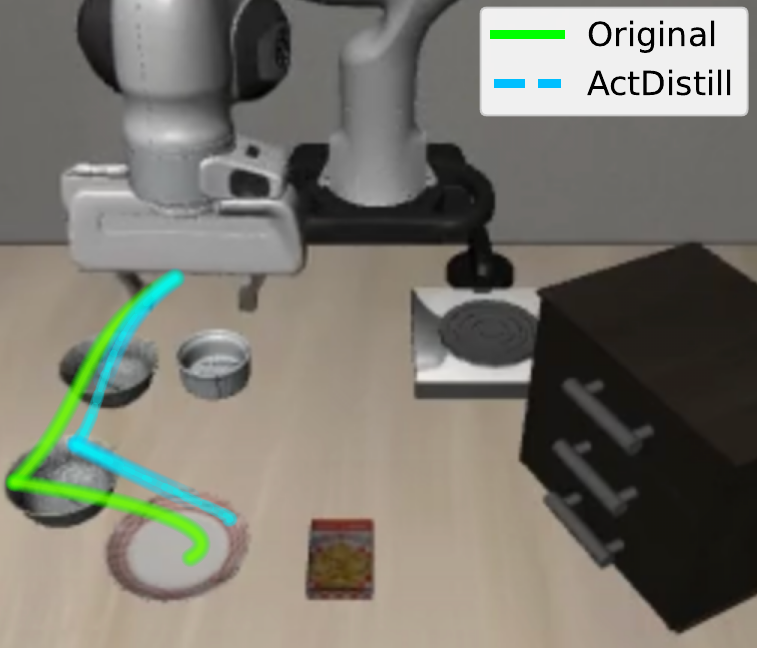}
        \caption{Instruction: Pick up the black bowl between the plate and the ramekin and place it on the plate.}
        \label{fig:traj_b}
    \end{subfigure}
    \caption{Comparison of manipulation trajectories between the original and ActDistill-optimized VLA models.}
    \label{fig:failures}
\end{figure}

\begin{figure}[t]
\centering
\begin{subfigure}[t]{\linewidth}
\centering
\includegraphics[width=0.98\linewidth]{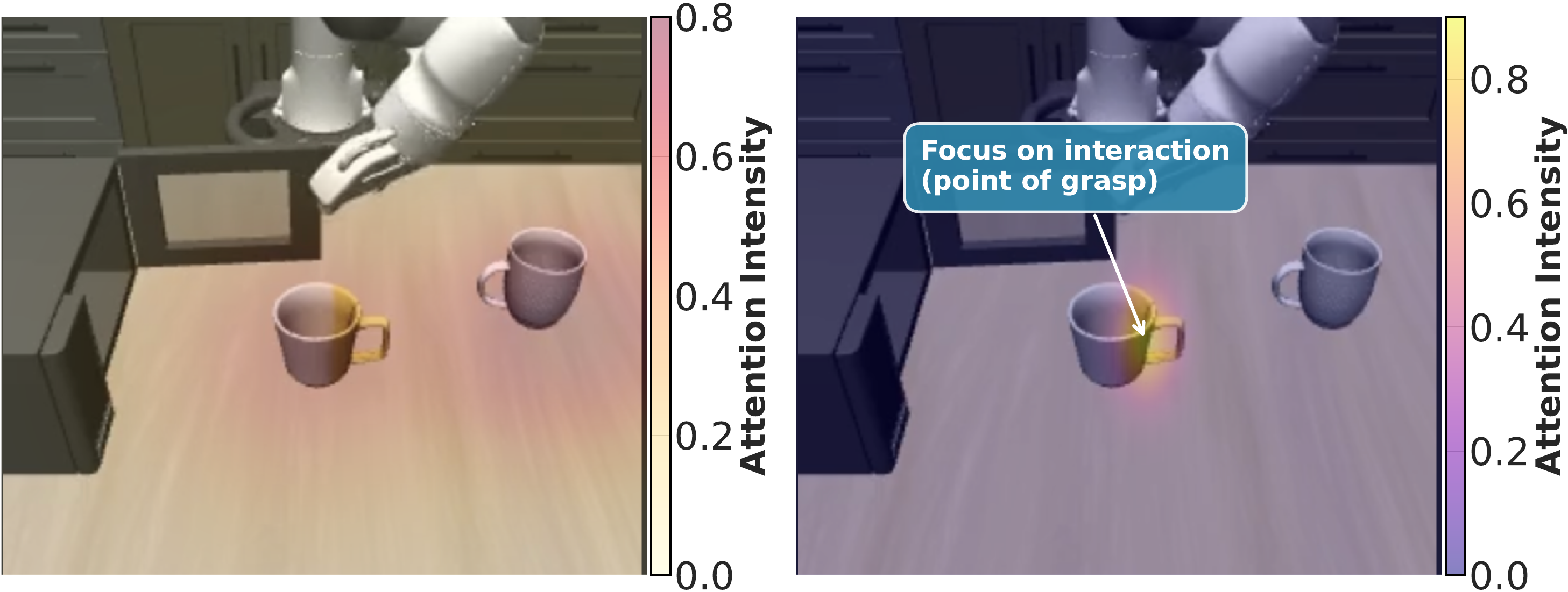}
\caption{Instruction: Put the cup on the left into the microwave.}
\end{subfigure}
%\vspace{4pt} % 控制上下间距，可调整或删除
\begin{subfigure}[t]{\linewidth}
\centering
\includegraphics[width=0.98\linewidth]{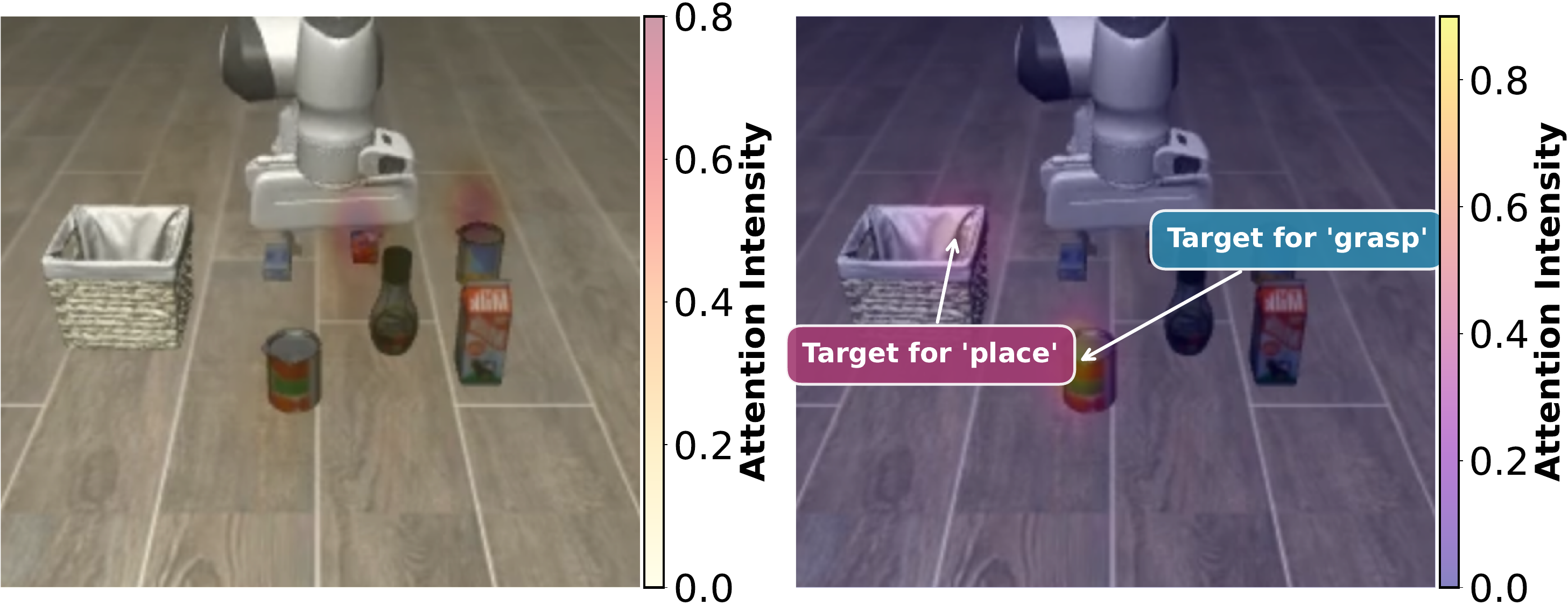}
\caption{Instruction: Put the red can into the basket.}
\end{subfigure}
\vspace{-0.5mm}
\caption{Final-layer attention heatmaps comparing the original and ActDistill-optimized VLA models, where the former attends broadly across objects while the latter concentrates on action-relevant regions.}
\label{fig:cam_vertical}
\end{figure}

\begin{figure*}[t]
\centering
\includegraphics[width=\linewidth, height=6.2cm]{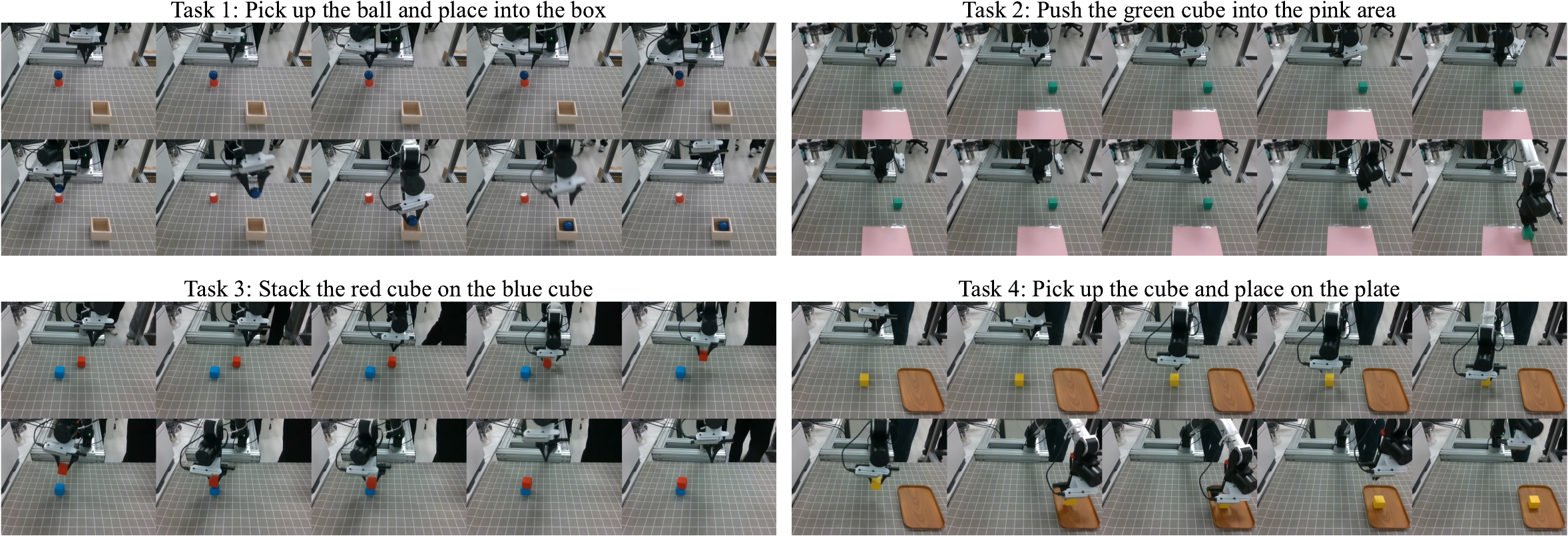}
\caption{Qualitative results of ActDistill in real-world robotic manipulation on ARX5 robot arm. The frame sequences illustrate the execution processes of four representative tasks where ActDistill generates smooth, accurate, and stable trajectories across diverse physical interactions.}
\label{fig:visualization1}
\end{figure*}

\subsection{More Exploration}
\label{sec:exploration}

\textbf{Dynamic Routing Intensity.}
We examine how routing strength affects the accuracy-efficiency trade-off by varying the number of skipped layers $n$ in the SIMPLER Visual Matching scenario. As shown in Figure~\ref{fig:skip_curve}, increasing $n$ steadily reduces FLOPs and latency, implying that many intermediate layers contribute little to the final decision. Moderate skipping preserves performance because action-guided routing bypasses redundant computations while preserving critical reasoning paths, whereas aggressive skipping degrades accuracy when layers essential for fine spatial or temporal refinement are omitted. The activation patterns in Figure~\ref{fig:visualization} confirm this trend: activations concentrate in early perceptual stages, and many mid-to-late layers are selectively deactivated. Unlike early-exit methods that stop inference once confident, ActDistill performs the full forward pass but suppresses unnecessary late-stage computations, reducing cost while retaining the layers needed for precise control without premature termination.

\begin{table}[!t]
\centering
\caption{Quantitative results on real-world manipulation tasks. Success rates are reported as percentages over 20 trials.}
\vspace{-1mm}
\label{tab:real_world}
\renewcommand{\arraystretch}{0.975}
\setlength{\tabcolsep}{5pt}
\resizebox{\linewidth}{!}{
\begin{tabular}{lcccccc}
\toprule
\multirow{2}{*}{Method} & \multicolumn{4}{c}{Success Rate (\%) $\uparrow$} & \multirow{2}{*}{Avg. $\uparrow$} & \multirow{2}{*}{Time (s) $\downarrow$} \\
\cmidrule(lr){2-5}
 & Task1 & Task2 & Task3 & Task4 & & \\
\midrule
Baseline           & 80\% & 65\% & 65\% & 40\% & 62.5\% & 10.2 \\
VLA-Cache          & 75\% & 55\% & 65\% & 25\% & 55.0\% & 8.7  \\
\textbf{ActDistill} & \textbf{80\%} & \textbf{60\%} & \textbf{75\%} & \textbf{35\%} & \textbf{62.5\%} & \textbf{6.3} \\
\bottomrule
\end{tabular}}
\end{table}

\vspace{1.5mm}
\noindent\textbf{Trajectory Comparison.}
Figure~\ref{fig:failures} visualizes robot trajectories on the LIBERO benchmark for both the full model and ActDistill. Both approaches produce smooth and accurate motions, indicating that ActDistill preserves trajectory stability even under accelerated execution. The trajectories further show that ActDistill captures object relationships and geometric constraints more effectively, enabling the manipulator to plan collision-free paths and achieve precise contact. These results suggest that action-guided semantics improve spatial understanding and lead to more reliable control in dynamic environments.

\vspace{1.5mm}
\noindent\textbf{Visualization of Action-Guided Attention.}
Figure~\ref{fig:cam_vertical} compares attention heatmaps extracted from the final layer of the baseline and ActDistill. The baseline, guided by general visual semantics, distributes attention across all visible objects, emphasizing category recognition. In contrast, ActDistill, driven by action-oriented semantics, concentrates on the precise interaction regions such as object handles or target containers. This focused pattern reflects a transition from general perception to functional reasoning, enabling the model to infer spatial relations and object geometry crucial for manipulation.

\subsection{Real-World Robot Experiments}
\label{sec:real_world}
To evaluate the practical deployment capability and robustness of ActDistill under real-world sensory noise and hardware latency, we conduct experiments on an ARX5 robotic arm across four representative manipulation tasks, as shown in Figure~\ref{fig:visualization1}. To assess policy robustness, we apply spatial randomization to the initial poses of all objects. We compare ActDistill against the full-scale OpenVLA baseline and the VLA-Cache method. As reported in Table~\ref{tab:real_world}, ActDistill achieves an average success rate of 62.5\%, matching the full-size OpenVLA model, while reducing the average inference time from 10.2\,s to 6.3\,s (a 1.62$\times$ speedup). In contrast, VLA-Cache exhibits a noticeable drop in performance, achieving only a 55.0\% average success rate. These results demonstrate that ActDistill maintains high task reliability under realistic conditions while significantly improving execution efficiency, making it suitable for latency-sensitive real-world robotic deployment.

\vspace{-2mm}
\section{Conclusion}
\label{sec:conclusion}
This paper presents ActDistill, a general action-guided self-derived distillation framework for efficient VLA models. The key insight is to align efficiency with the VL-to-Action transformation, rather than relying solely on vision-language redundancy. Accordingly, ActDistill transfers hierarchical action semantics from a teacher to a lightweight student and enables adaptive computation via action-aware routing. Experiments on both autoregressive and diffusion-based architectures show that ActDistill achieves comparable or superior performance while reducing computation by over 50\%, highlighting the effectiveness of action-centric modeling for efficient embodied intelligence.

\vspace{1.5mm}
\noindent\textbf{Limitation and Future Work.}
Although ActDistill exhibits strong efficiency and generality, it depends on pretrained teachers for structured supervision. Future work will explore teacher-free or reinforcement-guided variants that learn action priors autonomously and integrate long-horizon temporal reasoning into routing, advancing toward more adaptive and self-evolving embodied intelligence. 
%%
%% The acknowledgments section is defined using the "acks" environment
%% (and NOT an unnumbered section). This ensures the proper
%% identification of the section in the article metadata, and the
%% consistent spelling of the heading.
\begin{acks}
To Robert, for the bagels and explaining CMYK and color spaces.
\end{acks}

%%
%% The next two lines define the bibliography style to be used, and
%% the bibliography file.
\bibliographystyle{ACM-Reference-Format}
\bibliography{sample-base}

%%
%% If your work has an appendix, this is the place to put it.
\appendix

\clearpage
\setcounter{page}{1}

\section*{A. Theoretical Foundations}
We outline the theoretical basis of ActDistill by identifying redundancy in VLA computation, introducing the graph-based principle that supports efficient action prediction, and explaining why graph structures provide the inductive biases that Transformers alone cannot offer.

\subsection*{A.1 Computational Redundancy in VLA Models}
\label{sec:computational_redundancy}
In embodied manipulation, only a small subset of scene entities, such as the target object, its receptacle, and the relevant contact regions, directly determines the robot's control outcome, while most background tokens have negligible influence. Let the multimodal representations at layer $l$ be $\bm{H}_l = \{\bm{h}_{l,1}, \dots, \bm{h}_{l,N}\}$.
Empirical analyses across manipulation benchmarks show that the action mapping can be effectively characterized by a limited set of task-relevant features and their localized geometric relations:
\setcounter{equation}{12}
\begin{equation}
f(\bm{H}_l) \approx f(\{\bm{h}_{l,i}\}_{i \in S_l},\ \ \ \{(\bm{h}_{l,i}, \bm{h}_{l,j})\}_{(i,j) \in \mathcal{E}_l}),
\end{equation}
where $S_l$ indexes entities involved in the physical interaction and $\mathcal{E}_l$ captures their operational dependencies (e.g., grasp points, relative poses).

However, standard VLA backbones employ dense self-attention over all $N$ tokens, implicitly treating the scene as a complete interaction and incurring $\mathcal{O}(N^2)$ pairwise computations at every layer. This dense connectivity far exceeds the physically required structure for manipulation, producing substantial representational and computational redundancy across layers.

\subsection*{A.2 Graph-Based Efficiency Strategy in ActDistill}
To counter the dense $\mathcal{O}(N^2)$ interactions in standard VLA backbones, ActDistill restructures the teacher's representations into an explicitly sparse, action-aligned topology. Given the hidden states $\bm{H}_l = \{\bm{h}_{l,1}, \dots, \bm{h}_{l,N}\}$, a data-dependent affinity matrix is first computed, and only the most informative neighbors are retained:
\begin{equation} 
\bm{A}_l(i,j) = \frac{\hat{\bm{A}}_l(i,j)}{\sum_{j \in \mathrm{TopK}_k(i)} \hat{\bm{A}}_l(i,j)}, \ \ \ j \in \mathrm{TopK}_k(i). 
\end{equation}
Here, $\hat{\bm{A}}_l(i,j) = \exp(\phi(\bm{h}_{l,i})^\top \psi(\bm{h}_{l,j}))$ denotes the affinity scores, and $\phi$, $\psi$ are lightweight linear projections. By restricting attention to the top-$k$ neighbors of each token, ActDistill constructs a normalized sparse adjacency $\bm{A}_l$ that reflects the localized dependencies most predictive of manipulation behavior.

The resulting graph supports a compact message-passing refinement,
\begin{equation} 
\tilde{\bm{h}}_{l,i} = \sigma_r(\sum_{j \in \mathrm{TopK}_k(i)} \bm{A}_l(i,j)\,\bm{W}_l \bm{h}_{l,j}), 
\end{equation}
where $\bm{W}_l$ is a layer-specific transformation and $\sigma_r(\cdot)$ is a ReLU activation. The refined features are then aggregated through attention pooling to form a semantic capsule $\bm{s}_l^{\text{tea}}$. Driven by the distillation objective, this supervision steers the learned sparse topology toward the true action-relevant structure, ensuring that the encapsulated semantics remain aligned with the control signal.

In effect, ActDistill substitutes the backbone's dense, indiscriminate attention with a physically grounded sparse topology that reflects the action-critical dependencies of manipulation, thereby mitigating the computational redundancy highlighted above.

\subsection*{A.3 Limitations of Transformers for Efficient VLA Modeling}
While Transformers provide strong multimodal reasoning capabilities, their default computation pattern is misaligned with the structural properties of embodied manipulation. The self-attention mechanism processes all $N$ tokens through dense, all-to-all interactions, incurring substantial computational cost and coupling features that are not relevant to the control signal.

From a unified graph perspective, a Transformer layer can be viewed as a message passing network operating on a complete graph. Formally, its update corresponds to the GNN message passing
\begin{equation}
\bm{h}_{l+1, i} = \sigma_r( \sum_{j\in\mathcal{N}(i)} a_{l, ij}\bm{W}_{l}\bm{h}_{l,j})
\end{equation}
under the mapping $\mathcal{N}(i)=\{1,\dots,N\}$. Thus, Transformers assume full connectivity and soft, dense aggregation, whereas graph-based models exploit sparse topologies and localized neighborhoods for efficient reasoning.

Instead of relying on the Transformer's dense attention to implicitly learn which dependencies matter, ActDistill enforces an explicit sparse topology through Top-$k$ graph construction and action-aligned message passing. This contrasts with the Transformer's unconstrained attention, where irrelevant background tokens inevitably inject noise through a softmax mechanism, diluting the distinctiveness of task-critical features. By replacing dense aggregation with a structured, task-relevant neighborhood, ActDistill yields cleaner semantics and a high-fidelity supervision signal for distillation.

In summary, the limitation of Transformers lies not in their expressiveness, but in their inefficient fully connected computation graph. ActDistill addresses this mismatch by introducing a sparse, action-guided graph structure that better reflects the physical locality of manipulation and supports efficient VLA modeling.

\section*{B. Technical Specifications}
\subsection*{B.1 Model Architecture}
\noindent\textbf{Graph-Structured Encapsulation.}
To model local interactions within multimodal embeddings, we build $k$-nearest neighbor graphs with $k=8$ for all layers. Affinity projections $\varphi(\cdot)$ and $\psi(\cdot)$ map features from $\mathbb{R}^{4096}$ to $\mathbb{R}^{512}$ only for computing pairwise affinities and sparsification. Given the pre-activations $\bm{h}_{l,i}$, we compute $\hat{\bm{A}}_l(i,j)=\exp(\varphi(\bm{h}_{l,i})^\top \psi(\bm{h}_{l,j}))$ and retain the top-$k$ entries in each row. The selected rows are then $\ell_1$-normalized with a small $\epsilon=10^{-12}$ to ensure numerical stability. We then perform message passing and pooling on the original $4096$-d node features. Each graph is processed by a two-layer GAT. The attention pooling projection $\mathbf{w}_p\in\mathbb{R}^{4096}$ aggregates node information. The pooled feature is then transformed into a $512$-d capsule.
This mixed-dimensional design preserves rich per-token semantics while keeping graph construction compact and well-conditioned.

\vspace{1.5mm}
\noindent\textbf{Action Prediction Heads.}
Each layer is equipped with an auxiliary action head that provides per-layer action supervision, covering 7 degrees of freedom including 3D translation, 3D rotation, and the gripper command. These heads are used only during training and are removed at test time. Inference relies solely on the backbone's native policy head.

\vspace{1.5mm}
\noindent\textbf{Normalization and Initialization.}
All pooled embeddings are standardized using dataset-level statistics. Linear layers are initialized with Kaiming uniform initialization, while layer normalization uses unit scale and zero bias. Dropout is applied only to the GAT modules and auxiliary action heads.

\subsection*{B.2 Training Configuration}
\noindent\textbf{Stage I: Graph-Structured Encapsulation.}
We freeze the teacher VLA backbone and train only the graph encoders $\{\mathcal{G}_l\}$ together with their auxiliary action heads $\{\mathcal{H}_l^{\text{tea}}\}$. Training uses AdamW with a learning rate of $1\times10^{-6}$ and cosine annealing with a 1000-step warm-up, weight decay of $0.01$, and a batch size of $64$. The model is trained for one epoch on a single RTX 5090 for 4 hours. Mixed-precision training (bfloat16) is enabled, and gradient clipping with a max-norm of $1.0$ is applied. Teacher features are cached on the fly and remain frozen throughout, with no backpropagation through the teacher backbone.

\vspace{1.5mm}
\noindent\textbf{Stage II: Action-Guided Self-Derived Distillation.}
The student model and dynamic router are trained jointly in an end-to-end manner on RTX 5090 $\times$ 2 GPUs. We use AdamW with a learning rate of $1\times10^{-6}$ under cosine annealing with a 1000-step warm-up, a weight decay of $0.01$, and a global batch size of $128$ ($64$ per device). Training employs gradient clipping with a max-norm of $1.0$, bfloat16 mixed precision, and gradient checkpointing for memory efficiency, and runs for five epochs. The routing threshold selected via validation is $\tau=0.5$. Routing gates are initialized with low activation scores so that the student begins with a sparse execution pattern, then gradually activates a compact subset of layers sufficient for accurate action prediction, producing a computation path shaped directly by action demands.

\vspace{1.5mm}
\noindent\textbf{Losses and Layer Weights.}
The total loss integrates multiple objectives with weights $\alpha=1.0$, $\beta=1.0$, $\eta=0.5$ and uses load-balancing regularization with coefficient $\gamma=0.05$. Layer-wise weights follow a power-law:
\begin{equation} 
 \lambda_l=(\frac{l}{L})^{\kappa}, \ \ \  \kappa=2,
\end{equation} 
assigning larger weights to deeper layers to emphasize high-level semantics. We adopt cosine similarity for instance alignment and Frobenius norm for relation-level alignment.

\vspace{1.5mm}
\noindent\textbf{Action Model.}
We keep each backbone's native action head during training and inference: autoregressive decoding for OpenVLA, diffusion-based decoding for CogACT. ActDistill operates on backbone representations to learn the dynamic router, no change to the original policy head is required and auxiliary heads $\{\mathcal{H}_l\}$ are used as training probes.

\subsection*{B.3 Complete Pseudocode}
For completeness, we provide a more detailed pseudocode of the full ActDistill workflow as Algorithm \ref{alg:actdistill_full_appendix}. While the main paper presents a brief overview, the expanded version below outlines the key steps in teacher probing, graph encapsulation, student distillation, and routed inference in a clearer and more explicit form.

\setcounter{algorithm}{1}
\begin{algorithm}[!t]
\caption{The Complete Pseudocode of ActDistill}
\label{alg:actdistill_full_appendix}
\small
\begin{algorithmic}[2]
\Require Teacher $\mathcal{T}$, student $\mathcal{S}$, router $\mathcal{R}$, data $\mathcal{D}=\{(\bm{v},\bm{l},\bm{a})\}$, layers $1\ldots L$, KNN $k$, weights $\{\lambda_l\}$, $\alpha,\beta,\gamma,\eta$, threshold $\tau$

\vspace{2pt}
\Statex \textbf{Stage I: Graph-Structured Encapsulation}
\State Freeze $\mathcal{T}$; initialize graph params $\{\phi_l,\psi_l,\bm{W}_l,\bm{w}_{p,l}\}$ and heads $\{\mathcal{H}_l^{\text{tea}}\}$
\For{each mini-batch $(\bm{v},\bm{l},\bm{a}) \in \mathcal{D}$}
    \State $\bm{v}_e \gets \mathcal{E}_v(\bm{v})$, $\bm{l}_e \gets \mathcal{E}_l(\bm{l})$
    \State $\{\bm{h}_l^{\text{tea}}\}_{l=1}^L \gets \mathcal{B}(\bm{v}_e,\bm{l}_e)$
    \State $\mathcal{L}_{\text{aux}} \gets 0$
    \For{$l=1$ \textbf{to} $L$}
        \State Compute $\hat{\bm{A}}_l(i,j) = \exp(\phi_l(\bm{h}_{l,i}^{\text{tea}})^\top \psi_l(\bm{h}_{l,j}^{\text{tea}}))$
        \State For each $i$, keep $\text{TopK}_k(i)$ to get $\bm{A}_l(i,j)$
        \State $\tilde{\bm{h}}_{l,i} \gets \sigma_r\big(\sum_{j\in\text{TopK}_k(i)} \bm{A}_l(i,j)\bm{W}_l\bm{h}_{l,j}^{\text{tea}}\big)$
        \State $\alpha_{l,i} \gets \frac{\exp(\bm{w}_{p,l}^\top \tilde{\bm{h}}_{l,i})}{\sum_u \exp(\bm{w}_{p,l}^\top \tilde{\bm{h}}_{l,u})}$,\quad $\bm{s}_l^{\text{tea}} \gets \sum_i \alpha_{l,i}\tilde{\bm{h}}_{l,i}$
        \State $\hat{\bm{a}}_l^{\text{tea}} \gets \mathcal{H}_l^{\text{tea}}(\bm{s}_l^{\text{tea}})$
        \State $\mathcal{L}_{\text{aux}} \gets \mathcal{L}_{\text{aux}} + \|\hat{\bm{a}}_l^{\text{tea}}-\bm{a}\|_2^2$
    \EndFor
    \State Update $\{\phi_l,\psi_l,\bm{W}_l,\bm{w}_{p,l},\mathcal{H}_l^{\text{tea}}\}$ by $\nabla \mathcal{L}_{\text{aux}}$
\EndFor

\vspace{2pt}
\Statex \textbf{Stage II: Action-Guided Self-Derived Distillation}
\State Initialize $\mathcal{S}$, $\mathcal{R}$, student graph params $\{\phi_l^{\text{stu}},\psi_l^{\text{stu}},\bm{W}_l^{\text{stu}},\bm{w}_{p,l}^{\text{stu}}\}$ and heads $\{\mathcal{H}_l^{\text{stu}}\}$
\For{each mini-batch $(\bm{v},\bm{l},\bm{a}) \in \mathcal{D}$}
    \State $\bm{v}_e \gets \mathcal{E}_v(\bm{v})$, $\bm{l}_e \gets \mathcal{E}_l(\bm{l})$
    \State $g_l \gets \sigma_s(\bm{w}_{r,l}^\top[\bm{v}_e;\bm{l}_e])$ for $l=1\ldots L$
    \State $\bm{z}_0 \gets [\bm{v}_e;\bm{l}_e]$
    \For{$l=1$ \textbf{to} $L$}
        \State $\bm{z}_l^{\text{tmp}} \gets \mathcal{B}_{\text{stu}}^{(l)}(\bm{z}_{l-1})$
        \State $\bm{z}_l \gets g_l \bm{z}_l^{\text{tmp}} + (1-g_l)\bm{z}_{l-1}$
        \State Build KNN graph on tokens of $\bm{z}_l$ with ($\phi_l^{\text{stu}}$,$\psi_l^{\text{stu}}$, $\bm{W}_l^{\text{stu}}$,$\bm{w}_{p,l}^{\text{stu}}$) to get $\bm{s}_l^{\text{stu}}$
        \State $\hat{\bm{a}}_l^{\text{stu}} \gets \mathcal{H}_l^{\text{stu}}(\bm{s}_l^{\text{stu}})$
    \EndFor
    \State For each $l$, compute $\mathcal{L}_{\text{sem}}^{(l)}, \ \mathcal{L}_{\text{act}}^{(l)}$
    \State $\mathcal{L}_{\text{distill}} \gets \sum_{l=1}^{L}\lambda_l(\alpha\mathcal{L}_{\text{sem}}^{(l)}+\beta\mathcal{L}_{\text{act}}^{(l)})$
    \State $\bar{g} \gets \frac{1}{L}\sum_l g_l$, \quad $\mathcal{L}_{\text{lb}} \gets \sum_{l=1}^{L}(g_l-\bar{g})^2$
    \State $\mathcal{L}_{\text{total}} \gets \mathcal{L}_{\text{distill}} + \gamma \mathcal{L}_{\text{lb}}$
    \State Update $\mathcal{S}$, $\mathcal{R}$, $\{\mathcal{H}_l^{\text{stu}}\}$ by $\nabla \mathcal{L}_{\text{total}}$; keep teacher frozen
\EndFor

\vspace{2pt}
\Statex \textbf{Stage III: Inference}
\State Given $(\bm{v},\bm{l})$, compute $\bm{v}_e,\bm{l}_e$ and $g_l = \sigma_s(\bm{w}_{r,l}^\top[\bm{v}_e;\bm{l}_e])$
\State $\bm{z}_0 \gets [\bm{v}_e;\bm{l}_e]$, $L^* \gets 0$
\For{$l=1$ \textbf{to} $L$}
    \If{$g_l \ge \tau$}
        \State $\bm{z}_l \gets \mathcal{B}_{\text{stu}}^{(l)}(\bm{z}_{l-1})$
    \Else
        \State $\bm{z}_l \gets \bm{z}_{l-1}$
    \EndIf
\EndFor
\State Predict action $\hat{\bm{a}} \gets \mathcal{H}^{\text{stu}}(\bm{z}_{L^*})$
\end{algorithmic}
\vspace{-0.9mm}
\end{algorithm}

\section*{C. Evaluation Benchmarks}

\subsection*{C.1 LIBERO Task Suites}
We use the full set of task suites provided by LIBERO for evaluation and analysis. LIBERO is a manipulation benchmark built on \texttt{robosuite}, with procedurally generated scenes and object layouts. It organizes tasks into four suites that probe different aspects of transferable knowledge: \emph{LIBERO-Spatial}, \emph{LIBERO-Object}, \emph{LIBERO-Goal}, and \emph{LIBERO-Long}. The first three focus respectively on spatial relations, object semantics, and goal-level reasoning, while the long-horizon suite combines heterogeneous tasks to assess entangled transfer and extended sequential decision-making. Concretely, our evaluation follows the breakdown below:
\begin{itemize}
  \item \textbf{LIBERO-Spatial Suite.} Tasks focus on geometric relations and spatial rearrangement under layout randomization (e.g., drawer pose, receptacle placement). A representative instance is \emph{place bowl on plate with spatial variation}, which requires the policy to localize both the receptacle and the object under pose perturbations, execute a collision-free pick-and-place, and satisfy a final-on relation while maintaining a stable grasp and orientation. Spatial distractors and altered kinematic constraints test robustness to pose shifts and partial occlusions.
  \item \textbf{LIBERO-Object Suite.} Tasks center on object semantics and category transfer, such as grasping shapes or materials with distinct affordances. A typical instance is \emph{pick object}, where the target may be sampled from a set \{ketchup, bowl, apple, \dots\}. The agent must ground the linguistic category in the current scene, choose an appropriate grasp, and lift or transport the object to a designated location or above a height threshold. Object textures and clutter distributions are randomized per episode to promote semantic generalization.
  \item \textbf{LIBERO-Goal Suite.} Tasks emphasize instruction-to-effect grounding with composite goals, such as \emph{open target drawer; place apple into drawer}. The policy must correctly sequence subgoals (reach, grasp, actuate, stow), maintain state under partial observability (e.g., drawer state, object containment), and respect action preconditions. Goal predicates query both articulated-state targets and object relations.
  \item \textbf{LIBERO-Long Suite.} This suite collects temporally extended, multi-stage tasks that interleave articulation, relocation, and precise placement under distractors and occasional state resets. Episodes require consistent state tracking across phases (e.g., \emph{open} $\rightarrow$ \emph{insert} $\rightarrow$ \emph{close}), regrasping when necessary, and robust recovery from minor execution errors.
\end{itemize}

\begin{table*}[!t]
\centering
\setcounter{table}{3}
\caption{Ablation on graph neighbor size $k$ in the SIMPLER Visual Matching scenario.}
\label{tab:simpler_vm_graph_k}
\vspace{-1mm}
\renewcommand{\arraystretch}{0.975}
\setlength{\tabcolsep}{8.5pt}
\resizebox{\linewidth}{!}{
\begin{tabular}{lccccccccccc}
\toprule
\multirow{2}{*}{Variants} &  & & & \multicolumn{5}{c}{Success Rate ($\uparrow$)} & \multirow{2}{*}{Speed-up ($\uparrow$)} & \multirow{2}{*}{FLOPs ($\downarrow$)} \\
\cmidrule(lr){5-9}
 &  &  & & PickCan & MoveNear & Drawer & DrawerApple & Average &  &  \\
\midrule
{\color{gray} CogACT} & & & & {\color{gray}91.30\%} & {\color{gray}85.00\%} & {\color{gray}71.80\%} & {\color{gray}50.90\%} & {\color{gray}74.75\%} & {\color{gray}1.00$\times$} & {\color{gray}100.00\%} \\ \midrule
\rowcolor{gray!10}
\multicolumn{11}{l}{\textbf{ActDistill (default: $k{=}8$)}} \\
$k{=}4$ &  & & & 90.20\% & 81.10\% & 66.50\% & 46.80\% & 71.15\% & 1.68$\times$ & 41.90\% \\
$k{=}8$ & & & & 91.00\% & 82.30\% & 70.60\% & 52.40\% & 74.08\% & 1.67$\times$ & 42.30\% \\
$k{=}6$ & & & & 90.80\% & 82.10\% & 69.80\% & 51.70\% & 73.60\% & 1.66$\times$ & 42.80\% \\
$k{=}10$ & & & & 90.70\% & 81.90\% & 69.50\% & 51.50\% & 73.40\% & 1.66$\times$ & 42.60\% \\
\bottomrule
\end{tabular}}
\end{table*}

\begin{table*}[t]
\centering
\caption{Ablation on loss weight ratio $\alpha{:}\beta$ (semantic vs. action) in the SIMPLER Visual Matching scenario.}
\label{tab:simpler_vm_loss_ratio}
\vspace{-1mm}
\renewcommand{\arraystretch}{0.975}
\setlength{\tabcolsep}{8.5pt}
\resizebox{\linewidth}{!}{
\begin{tabular}{lcccccccccc}
\toprule
\multirow{2}{*}{Variants} & & & \multicolumn{5}{c}{Success Rate ($\uparrow$)} & \multirow{2}{*}{Speed-up ($\uparrow$)} & \multirow{2}{*}{FLOPs ($\downarrow$)} \\
\cmidrule(lr){4-8}
 & & & PickCan & MoveNear & Drawer & DrawerApple & Average &  &  \\
\midrule
{\color{gray} CogACT} & & & {\color{gray}91.30\%} & {\color{gray}85.00\%} & {\color{gray}71.80\%} & {\color{gray}50.90\%} & {\color{gray}74.75\%} & {\color{gray}1.00$\times$} & {\color{gray}100.00\%} \\
\midrule
\rowcolor{gray!10}
\multicolumn{10}{l}{\textbf{ActDistill (default: $\alpha{:}\beta{=}(1,1)$)}} \\
$\alpha{:}\beta{=}(1,0.5)$ & & & 90.50\% & 81.80\% & 68.50\% & 49.80\% & 72.65\% & 1.62$\times$ & 44.80\% \\
$\alpha{:}\beta{=}(1,1)$ & & & 91.00\% & 82.30\% & 70.60\% & 52.40\% & 74.08\% & 1.67$\times$ & 42.30\% \\
$\alpha{:}\beta{=}(1,2)$ & & & 90.90\% & 81.90\% & 69.60\% & 50.70\% & 73.28\% & 1.65$\times$ & 43.50\% \\
$\alpha{:}\beta{=}(3,4)$ & & & 90.20\% & 81.30\% & 69.90\% & 50.70\% & 73.03\% & 1.66$\times$ & 43.40\% \\
\bottomrule
\end{tabular}}
\end{table*}

\subsection*{C.2 SIMPLER (Google Robot Setup)}
We evaluate all task variants provided by the SIMPLER benchmark, focusing on the standardized \emph{Google Robot} (GR) setup and reporting results under both protocols. SIMPLER provides simulated counterparts to real-robot setups, with task templates instantiated in \texttt{ManiSkill2}/SAPIEN environments and consistent success predicates to enable reproducible real-to-sim correlation. Our experiments include the following tasks:
\begin{itemize}
  \item \textbf{Pick coke can.} Detect and grasp an opened coke can, then lift it above a height threshold or place it in a designated zone. In simulation, mass and friction parameters may be perturbed to test the robustness of the grasp strategy.
  \item \textbf{Move obj1 near obj2.} Relocate \emph{obj1} to be within a proximity band around \emph{obj2} while avoiding collisions and unintended state changes. Scenes randomize object identities and initial placements, encouraging relational grounding beyond category recognition.
  \item \textbf{Open/close top/middle/bottom drawer.} Three-level drawer articulations (top/middle/bottom) with variation in drawer geometry, friction, and handle pose. Tasks require reliable handle localization, stable pulling or pushing along the joint axis, and termination within the required joint-range band for success.
  \item \textbf{Open top drawer; place apple into top drawer.} A composite manipulation that first changes the drawer articulation and then performs precise placement (containment). The agent must maintain the object pose after any regrasp, avoid collisions with the drawer front and rails, and satisfy containment predicates at evaluation.
\end{itemize}

To ensure consistent and comprehensive evaluation, SIMPLER employs two complementary modes that capture performance under matched real-world conditions and robustness across task variations:
\begin{itemize}
\item \textbf{Visual Matching (VM)} compares policy behavior on simulated scenes that closely replicate real-world episodes in textures, lighting, and object poses. Environment seeds are curated to mirror the real layouts, and success is evaluated per episode and averaged over the matched set.
\item \textbf{Variant Aggregation (VA)} evaluates policies across a distribution of task variants, such as object orientations (horizontal/vertical/standing) for \emph{pick coke can} or articulation levels (top/middle/bottom) for drawers, aggregating success across sub-variants to reflect robustness beyond a single matched scene. 
\end{itemize}

SIMPLER provides canonical environment names aligned with the GR setup and exposes episode-level success predicates (lift height, proximity bands, articulation thresholds, and containment) for consistent measurement. Unless otherwise noted, we report mean success over fixed seeds for VM and VA, and additionally comment on latency and FLOPs when comparing efficiency methods.

\begin{table*}[t]
\centering
\caption{Ablation on routing threshold $\tau$ in the SIMPLER Visual Matching scenario.}
\label{tab:simpler_vm_routing_tau}
\vspace{-1mm}
\renewcommand{\arraystretch}{0.975}
\setlength{\tabcolsep}{8.5pt}
\resizebox{\linewidth}{!}{
\begin{tabular}{lccccccccccc}
\toprule
\multirow{2}{*}{Variants} & & & & \multicolumn{5}{c}{Success Rate ($\uparrow$)} & \multirow{2}{*}{Speed-up ($\uparrow$)} & \multirow{2}{*}{FLOPs ($\downarrow$)} \\
\cmidrule(lr){5-9}
 & & & & PickCan & MoveNear & Drawer & DrawerApple & Average &  &  \\
\midrule
{\color{gray} CogACT} & & & & {\color{gray}91.30\%} & {\color{gray}85.00\%} & {\color{gray}71.80\%} & {\color{gray}50.90\%} & {\color{gray}74.75\%} & {\color{gray}1.00$\times$} & {\color{gray}100.00\%} \\ \midrule
\rowcolor{gray!10}
\multicolumn{11}{l}{\textbf{ActDistill (default: $\tau{=}0.5$)}} \\
$\tau{=}0.4$ & & & & 91.20\% & 82.50\% & 70.90\% & 53.00\% & 74.40\% & 1.47$\times$ & 52.70\% \\
$\tau{=}0.5$ & & & & 91.00\% & 82.30\% & 70.60\% & 52.40\% & 74.08\% & 1.67$\times$ & 42.30\% \\
$\tau{=}0.6$ & & & & 90.70\% & 81.70\% & 69.30\% & 51.10\% & 73.20\% & 1.76$\times$ & 38.10\% \\
$\tau{=}0.7$ & & & & 89.90\% & 80.90\% & 67.40\% & 47.80\% & 71.50\% & 1.88$\times$ & 32.60\% \\
\bottomrule
\end{tabular}}
\end{table*}

\begin{figure*}[t]
\centering
% ---------- Top Row ----------
\begin{subfigure}{0.48\linewidth}
    \centering
    \includegraphics[width=\linewidth]{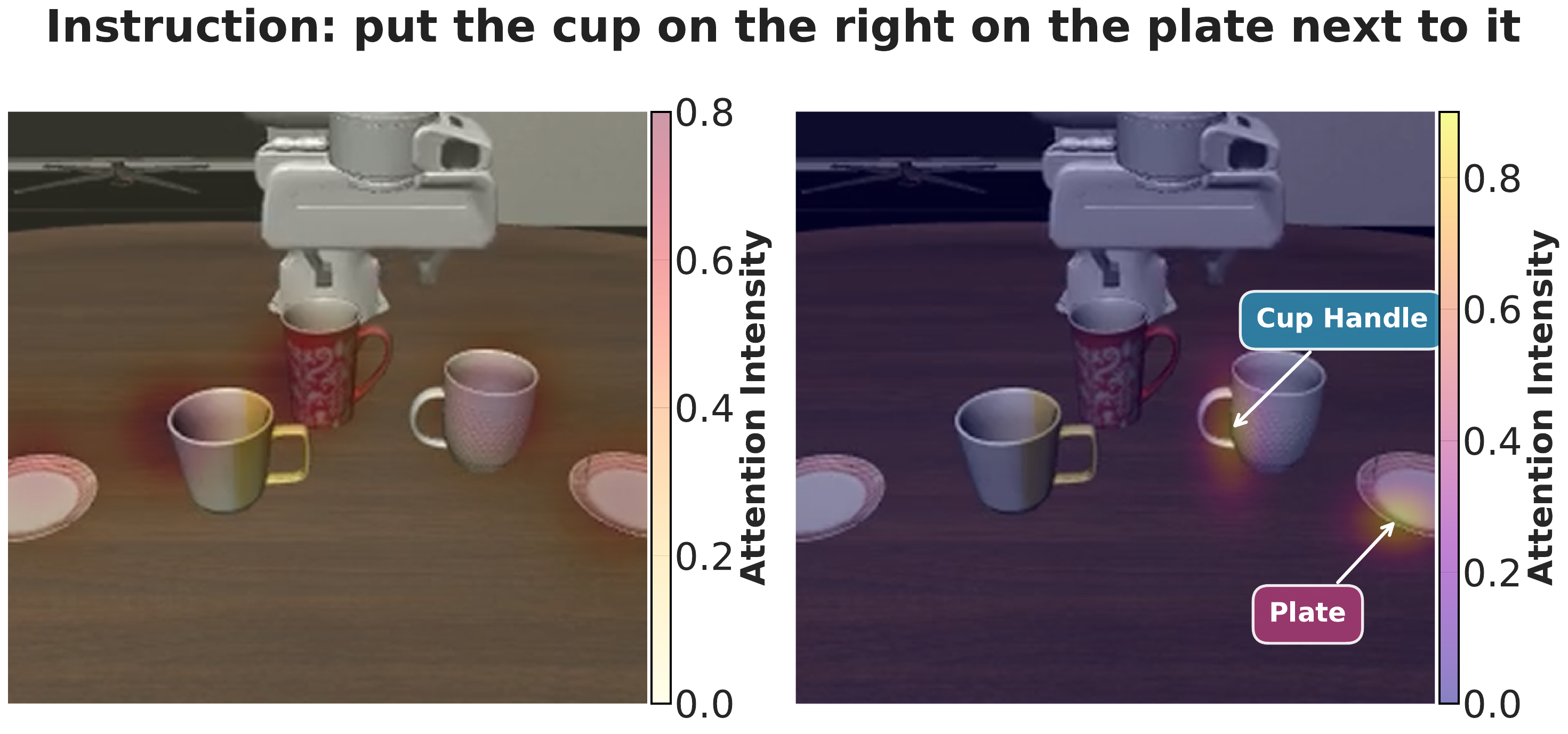}
    \label{fig:abcd_a}
\end{subfigure}
\hfill
\begin{subfigure}{0.49\linewidth}
    \centering
    \includegraphics[width=\linewidth]{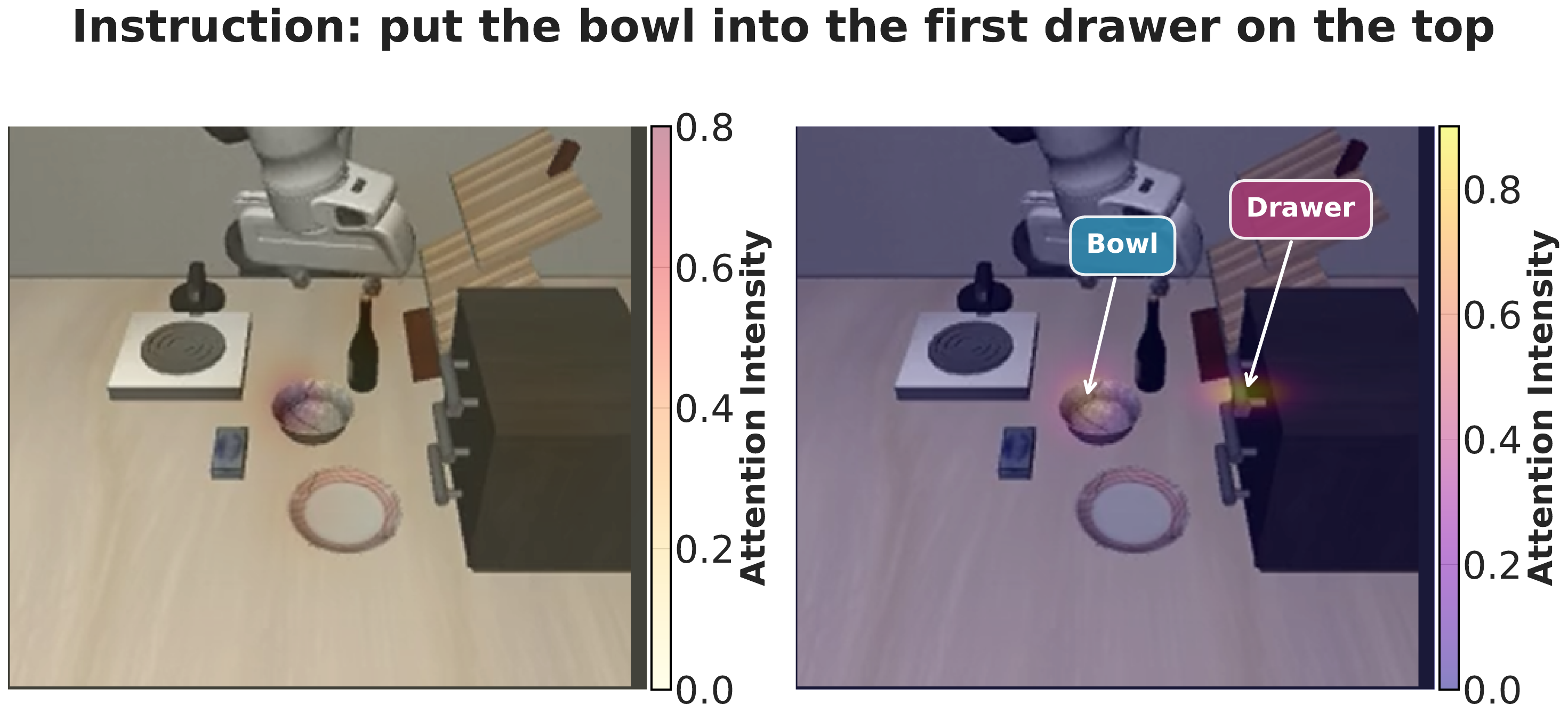}
    \label{fig:abcd_b}
\end{subfigure}
\vspace{0.2cm} % spacing between rows
% ---------- Bottom Row ----------
\begin{subfigure}{0.49\linewidth}
    \centering
    \includegraphics[width=\linewidth]{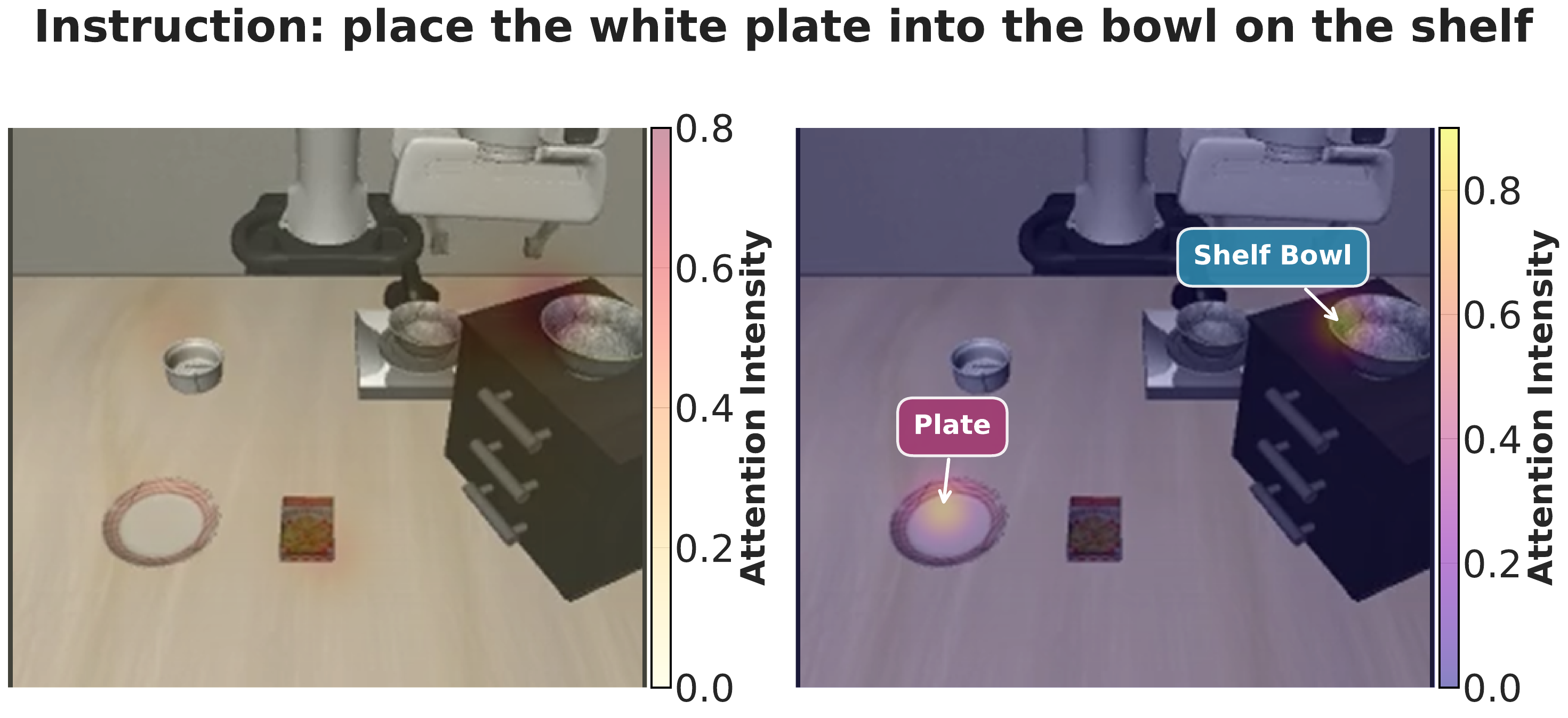}
    \label{fig:abcd_c}
\end{subfigure}
\hfill
\begin{subfigure}{0.49\linewidth}
    \centering
    \includegraphics[width=\linewidth]{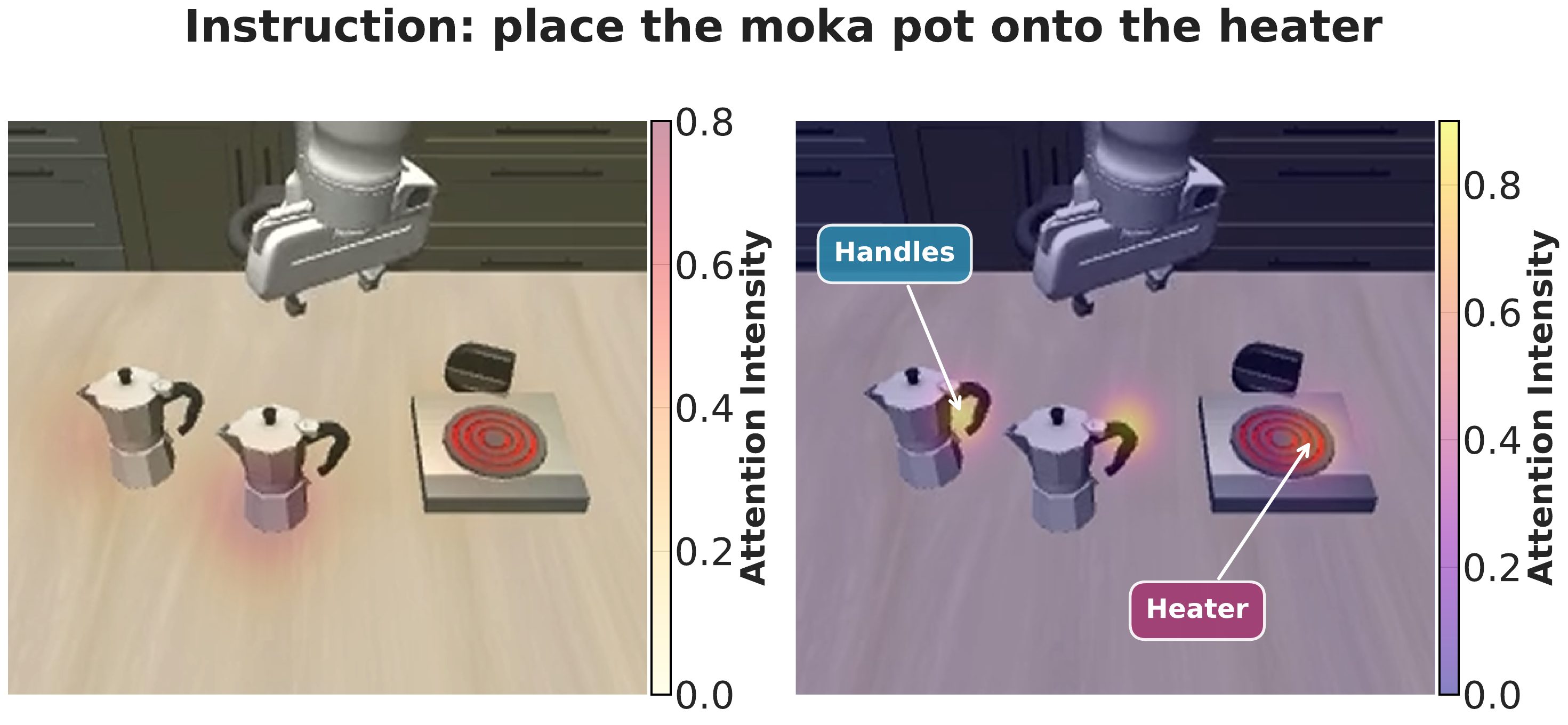}
    \label{fig:abcd_d}
\end{subfigure}
\vspace{-6mm}
\setcounter{figure}{7}
\caption{Additional heatmap visualizations comparing ActDistill and the teacher model.}
\label{fig:abcd}
\end{figure*}

\section*{D. Additional Experiments and Analysis}
\subsection*{D.1 Hyperparameter Ablations}
\subsubsection*{D.1.1 Graph Hyperparameter Ablations}
\label{sec:abl_k}
We vary the $k$-nearest neighbor size used in the graph-structured encapsulation and report SIMPLER-Visual Matching results in Table~\ref{tab:simpler_vm_graph_k}. Note that the graph modules are removed at inference, so test-time computation is determined solely by the learned routing rather than the graph construction itself.

\vspace{1.5mm}
\noindent\textbf{Results.}
We observe a shallow-\emph{U} trend: when $k$ is too small, relational cues are under-captured (for example, $k{=}4$ yields an average of $71.15\%$), whereas larger values offer only marginal improvement over the default (for example, $k{=}10$ reaches $73.40\%$). The default $k{=}8$ provides the best overall balance at $74.08\%$, with notable robustness on precision-sensitive tasks such as \textit{Drawer} and \textit{DrawerApple}. As expected, the Speed-up and FLOPs vary only slightly across different $k$ values (within $\pm0.02\times$ and $\pm0.9\%$), since the dynamic router determines which layers are executed at inference time.

\vspace{1.5mm}
\noindent\textbf{Takeaway.}
Relational sparsity controlled by a moderate $k$ (here, $k{=}8$) yields the most reliable action semantics under distillation, while inference efficiency remains largely unaffected by $k$ because routing is learned and graph modules are not used at test time.

\subsubsection*{D.1.2 Loss Weight Sensitivity}
\label{sec:abl_alpha_beta}
We vary the semantic-to-action loss ratio $\alpha{:}\beta$ in Eq.~(9) while keeping other settings unchanged, and summarize results in Table~\ref{tab:simpler_vm_loss_ratio}.

\vspace{1.5mm}
\noindent\textbf{Results.}
A balanced weighting of $\alpha{:}\beta = (1,1)$ yields the strongest average performance ($74.08\%$) with high efficiency (Speed-up $1.67\times$, $42.30\%$ FLOPs). Reducing the action term to $\alpha{:}\beta = (1,0.5)$ weakens fine control and lowers the average to $72.65\%$ with a speed-up of $1.62\times$. Increasing the action weight to $\alpha{:}\beta = (1,2)$ slightly disrupts relational alignment, producing an average of $73.28\%$ at $1.65\times$. These trends align with the component ablations in the main paper, where removing either semantic or action-level supervision causes larger performance drops.

\vspace{1.5mm}
\noindent\textbf{Takeaway.}
Balanced semantic and action guidance best preserves action-centric semantics and low-level precision under the same routing budget, reinforcing the complementary roles of the two losses.

\subsubsection*{D.1.3 Routing Threshold Analysis}
\label{sec:abl_tau}
We sweep the routing threshold $\tau$ that binarizes layer gates at the inference process, thereby directly controlling sparsity and overall computation. Results are reported in Table~\ref{tab:simpler_vm_routing_tau}. 

\vspace{1.5mm}
\noindent\textbf{Results.}
Lowering $\tau$ activates more layers, yielding higher FLOPs and lower speed (e.g., $\tau{=}0.4$: Speed-up $1.47\times$, $52.70\%$ FLOPs) with a slight gain in average success ($74.40\%$).
Raising $\tau$ induces more skipping and improves speed (e.g., $\tau{=}0.7$: Speed-up $1.88\times$, $32.60\%$ FLOPs) but reduces success ($71.50\%$), with the drop concentrated on precision-sensitive tasks (\textit{Drawer} $70.90\!\rightarrow\!67.40$, \textit{DrawerApple} $53.00\!\rightarrow\!47.80$ as $\tau$ increases from $0.4$ to $0.7$).
The default $\tau{=}0.5$ offers a favorable trade-off ($74.08\%$, Speed-up $1.67\times$, $42.30\%$ FLOPs), matching the performance-efficiency knee point observed in the main-text trade-off curve and activation frequency analysis.

\vspace{1.5mm}
\noindent\textbf{Takeaway.}
The router exhibits the expected accuracy-efficiency trade-off as $\tau$ varies, and a mid-range threshold offers robust control with substantial computational savings, consistent with the fact that inference-time execution is determined by gate discretization.

\subsection*{D.2 Additional Heatmap Visualizations}
To complement the qualitative analysis in the main paper, we provide additional heatmap
visualizations in Figure~\ref{fig:abcd}. These visualizations illustrate how ActDistill preserves
the teacher’s action-centric semantics while using a substantially more compact routed
computation pattern. Across diverse scenes, including cluttered environments, multi-object
layouts, and articulated settings, the distilled student consistently attends to the relevant
objects, contact regions, and geometric affordances required for manipulation. This behavior
mirrors the teacher’s high-level semantic structure and indicates that our action-guided
distillation effectively aligns per-layer capsules even under aggressive compute reduction.

\section*{E. Failure Analysis}
To better understand the limitations of ActDistill, we qualitatively analyze failure cases. We find that most errors fall into two dominant categories, while high-level action semantics remain largely intact.

\vspace{1.5mm}
\noindent\textbf{Teacher-Inherited Failures.}
Since ActDistill explicitly distills the teacher's action semantics, it naturally inherits systematic mistakes made by the teacher policy. Typical patterns include ambiguous language instructions or visually cluttered scenes where the teacher already fails to complete the task (e.g., stopping short of the required articulation range, or terminating with the object slightly outside the target receptacle). In these cases, the student closely tracks the teacher's behavior: it attends to the correct regions and follows a similar trajectory, but reproduces the same suboptimal stopping condition or contact strategy. While our action-guided distillation occasionally smooths trajectories and slightly improves stability, it fails to fundamentally correct such upstream policy biases.

\vspace{1.5mm}
\noindent\textbf{Insufficient Action Precision under Correct Semantics.}
A second class of failures arises when the student correctly identifies \emph{what} to interact with and \emph{where} to act, but executes with insufficient low-level precision to satisfy the benchmark success predicates. For example, the gripper may approach the correct object and contact region, but a slightly shallow or deep grasp offset causes the object to slip during lifting. Similarly, a drawer may be nearly closed yet fail because it stops just outside the required articulation range. Such errors occur more frequently in tasks with tight geometric or articulation tolerances, where small deviations in pose or timing can directly flip the binary success label.

In general, we rarely observe failures that correspond to semantic misinterpretation of the instruction, such as picking up a wrong object category, moving toward an entirely wrong receptacle, or targeting an incorrect interaction point on the scene geometry. Across failure episodes, attention maps and trajectories typically focus on the appropriate objects and functional regions. When the policy fails, it is usually due to inherited teacher behavior or control imprecision rather than incorrect action semantics. This supports our claim that action-guided distillation effectively preserves semantic grounding while shifting the primary bottlenecks to teacher quality and fine-grained control accuracy.
\end{document}